\documentclass[conference]{IEEEtran}
\usepackage{graphicx}
\usepackage{epstopdf} 
\usepackage[symbol]{footmisc}
 
\usepackage[numbers]{natbib}
\usepackage{amsmath}
\usepackage[font=scriptsize]{subcaption}
\usepackage[font=scriptsize]{caption}
\usepackage{xcolor}
\usepackage{multicol}
\usepackage[bookmarks=true]{hyperref}

\begin{document}

\title{Influence of Static and Dynamic Downwash Interactions on Multi-Quadrotor Systems}

\author{Anoop Kiran, Nora Ayanian, and Kenneth Breuer\\School of Engineering, Brown University, 
Providence, Rhode Island, USA\\Email: {\tt\footnotesize \{anoop\_kiran, nora\_ayanian, kenneth\_breuer\}@brown.edu}}

\maketitle

\begin{abstract}
Flying multiple quadrotors in close proximity presents a significant challenge due to complex aerodynamic interactions, particularly \emph{downwash effects} that are known to destabilize vehicles and degrade performance. Traditionally, multi-quadrotor systems rely on conservative strategies, such as collision avoidance zones around the robot volume, to circumvent this effect. This restricts their capabilities by requiring a large volume for the operation of a multi-quadrotor system, limiting their applicability in dense environments. 
This work provides a comprehensive, data-driven analysis of the downwash effect, with a focus on characterizing, analyzing, and understanding forces, moments, and velocities in both single and multi-quadrotor configurations. 
We use measurements of forces and torques to characterize vehicle interactions and particle image velocimetry (PIV) to quantify the spatial features of the downwash wake for a single quadrotor and an interacting pair of quadrotors. 
This data can be used to inform physics-based strategies for coordination, leverage downwash for optimized formations, expand the envelope of operation, and improve the robustness of multi-quadrotor control. 
\end{abstract}

\IEEEpeerreviewmaketitle 

\section{Introduction}
 
Quadrotor teams have found widespread use in numerous applications, including search and rescue~\cite{SchedlSearchRescue, HayatSAR}, precision agriculture~\cite{LottesUAVAgriculture, GrenwoodSprayWake}, and infrastructure inspection~\cite{NathanInfrastructurecInspection}. However, a critical weakness lies in the inability to operate dense formations of quadrotors due to the aerodynamic interference of rotor wakes, posing a significant challenge in advancing multi-robot aerial systems. Although this  ``downwash effect'' has been recognized in helicopter aerodynamics~\cite{LeishmanHeli}, research into its impact on small-scale aerial vehicles, such as quadrotors, is still in its infancy. Accurate analysis and understanding of this effect can extend use cases for efficient close proximity applications such as aerial docking~\cite{DongAerialLanding}, large-scale trajectory planning of quadrotor teams~\cite{HonigDownwashTraj, PanLargeScale}, sensor placement~\cite{GraffSensorPlacement}, and collaborative transport of objects~\cite{CotsakisCollabTransp} that currently avoid tight formation flight due to the unacceptable risks in maintaining stable flight. 

Detailed characterization of downwash effects, including forces, moments, velocity fields, and unsteady measurements, establishes a foundational empirical basis for aerodynamic interactions that can be used in physics-based modeling and learning-based approaches, essential for the future development of more sophisticated control strategies. Incorporating physics-based knowledge into data-driven methods has been proven to enable more sample-efficient learning~\cite{SmithDownwashFlight, CheeQuadrotorsTightFormations}, requiring fewer real-world trials to achieve reliable performance for dense formation flight. 
  
\begin{figure}[tbh]
    \centering    \includegraphics[width=0.47\textwidth]{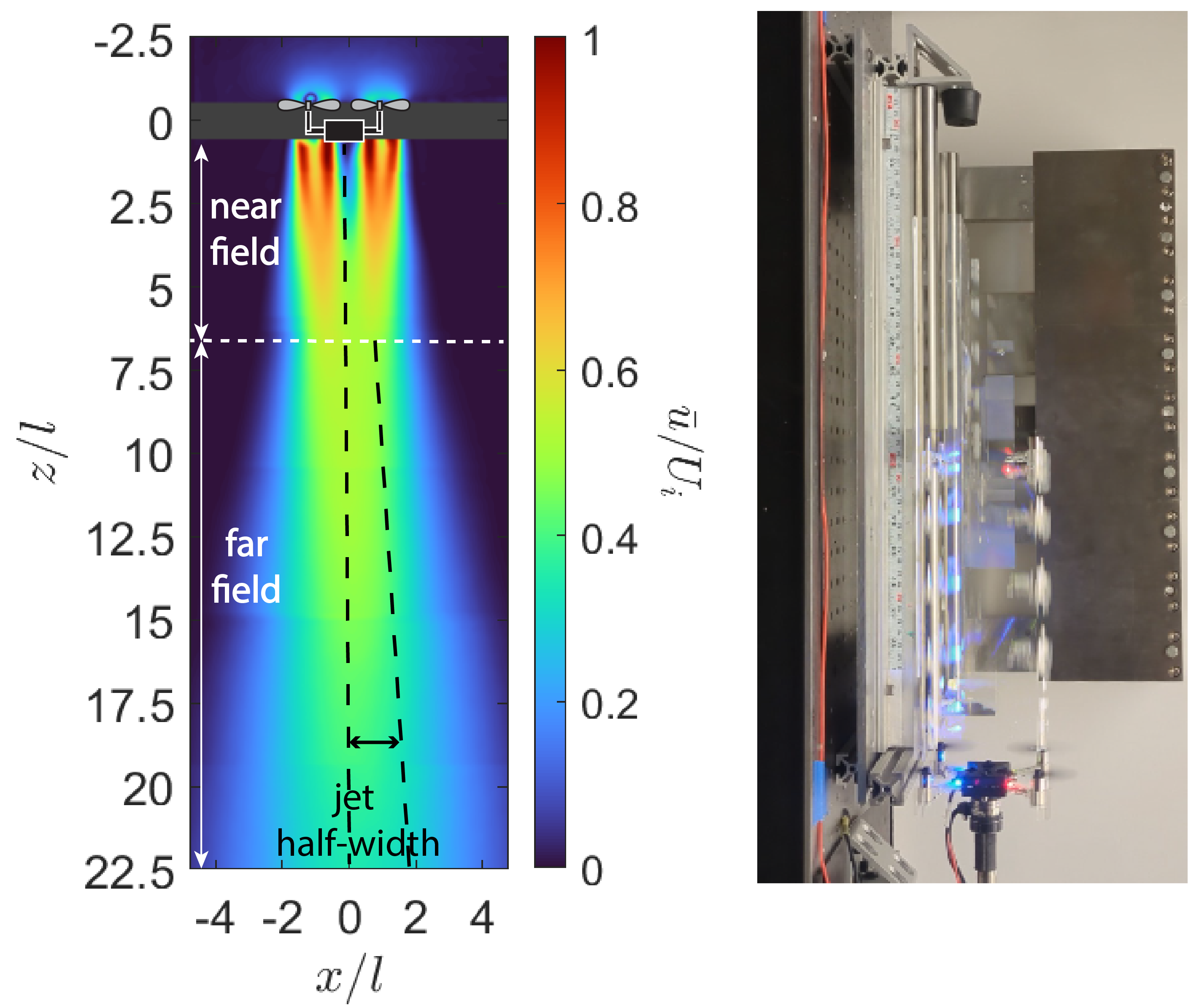}
    \caption{Axial downwash velocity, $\bar{u}$, (left) of downwash below a Crazyflie quadrotor. The individual propeller flows observed in the near-field merge to form a turbulent jet in the far-field ($z/l>6.5$). A long-exposure image (right) of dynamic interaction between quadrotors, where two hovering Crazyflies, one accelerated using a linear traverse (\emph{CrazyRail}) rapidly approaches another mounted on a load cell measuring forces and moments due to the interaction.
    \label{fig:cover_image}}
\end{figure}

\subsection{Contributions}
The contributions of this paper are twofold. Firstly, we characterize the effect of downwash on forces, moments, and velocities acting on a pair of quadrotors flying in close proximity as a function of their relative separation. Quantifying these critical variables provides a foundation for understanding the magnitudes of destabilizing effects in dense formations, aiding in model development for control. The influence of the separation between quadrotors and its subsequent impact on downwash is emphasized. We develop an \textit{algebraic reduced-order model} of quadrotor downwash that exhibits the characteristic scaling behavior of canonical turbulent round jets. Further characterization and analyses detailed herein provide a valuable open-access dataset\footnote{Dataset repository: \href{https://doi.org/10.26300/64d4-wa17}{https://doi.org/10.26300/64d4-wa17}} for the multi-robot community, which has traditionally addressed this challenge by avoiding flying within zones where downwash influence is dominant~\cite{NathanGraspTestbed, HonigDownwashTraj, PanLargeScale, DongAerialLanding}. 

Secondly, we quantify the unsteadiness (turbulence) due to downwash, helping to better understand destabilizing effects. Recognizing unsteady effects enables proactive corrective measures to maintain stability under the influence of downwash, optimize path planning, and control for dense formation flight.

\subsection{Related Work}
Recently, studying and mitigating the effects of the downwash from quadrotors has gained traction due to its critical impact on the stability, performance, and efficiency of quadrotor systems~\cite{GielisModelingDownwash, SmithDownwashFlight, BauersfeldRoboticsMeetsFluidDynamics}. Previous work has considered downwash in collision avoidance by modeling quadrotor volume as cylinders~\cite{NathanGraspTestbed, FerreraCollisionAvoidance} or axially-aligned ellipsoids~\cite{HonigDownwashTraj, BareissCollisionAvoidance, ArulCollisionAvoidance}, with a more extensive range along the axial direction of downwash influence. For collision avoidance, Preiss et al. used a vertical separation, $\Delta{z}$ = 60 cm (${{\Delta}z}/l \approx 18.5$, where $l$ is a characteristic scaling length, defined as the distance between the vehicle and rotor centers) and a horizontal separation of $\Delta{x}$ = 24 cm (${{\Delta}x}/l \approx 7.4$) between two small quadrotors (Crazyflies~\cite{Crazyflie}) in their experiments~\cite{HonigDownwashTraj}. However, these models employ predefined arbitrary separation between quadrotors, without factoring in flight physics. Therefore, a critical gap remains in understanding the spatial influence of downwash on quadrotor interactions at a fundamental level. 

Models based on first principles provide a foundation for quantifying established downstream flow profiles but struggle to capture the full range of aerodynamic effects, particularly in both the near-field and far-field regions~\cite{YeoEmpiricalModel, JainDownwashModel}. Motivated by the interaction between quadrotors, Jain et al. measured the decay of the rotor wake below a large, and a small quadrotor~\cite{JainDownwashModel}, and Yeo et al. measured the decay of the downwash below varying scales of aerial vehicles with propeller diameters ranging from small quadrotors, $l \sim 20$ cm to helicopters, $l \sim 50$ cm~\cite{YeoEmpiricalModel}, fitting parameters to a Gaussian velocity decay profile. In both these studies, anemometers were placed at an arbitrary downstream distance, assuming that the wake velocity had a Gaussian profile. While numerical simulations have demonstrated their ability to capture complex downwash wake structure~\cite{YoonComputationalAerodynamicModeling}, they are primarily limited in their flow field area due to computational complexity. However, the effect of downwash and its subsequent wake is strongly influenced by the relative proximity between quadrotors, necessitating accurate modeling of downwash for a comprehensive mapping both in the near-field and far-field regions for dense formations~\cite{CheeQuadrotorsTightFormations, SmithDownwashFlight}.

Moreover, these studies~\cite{YeoEmpiricalModel, JainDownwashModel, GielisModelingDownwash} report results in dimensional units for specific quadrotors, making it difficult to scale these findings to other vehicles or to reproduce the data in different contexts. Although it is possible to normalize other results, doing so retroactively can be nontrivial without detailed information on the custom designs or operating parameters of the vehicles. Hence, it is unclear if these velocity measurements, even in the far-field can generalize to a range of quadrotors of varying scales and sizes. Our findings address this gap by systematically characterizing and analyzing these normalized flow variables using Crazyflie quadrotors, which are widely adopted for research and experimental applications. 
 
Most recently, Bauersfeld et al. explored interesting quadrotor wake structure, including the merging of rotor jets, but relied on a low-bandwidth, relatively large velocity
probe to measure the wake structure below a range of commercial quadrotors~\cite{BauersfeldRoboticsMeetsFluidDynamics}. In this work, we employ particle image velocimetry (PIV)~\cite{MarkusPIV}, a measurement technique that captures spatially resolved flow fields. PIV provides finer details inaccessible to lower-resolution counterparts such as anemometers, which provide point measurements that must be extrapolated for spatial insights, compromising high-fidelity aerodynamic modeling. Our findings offer insights critical for validating and extending canonical models on turbulent jets~\cite{PopeTurbulentFlows} for quadrotor downwash wakes. 

While existing work has targeted avoiding the region of downwash due to destabilizing effects, this paper aims to understand and characterize these interactions, paving the way for quadrotors to operate within this region rather than avoid it. In addition to providing a deeper understanding of this aerodynamic phenomenon, this work lays the groundwork for developing robust control strategies to increase the flight envelope and performance in close-proximity multi-quadrotor systems.

\section{Methodology}

Consider a pair of quadrotors flying in close proximity, one below the other. Our approach aims to quantify the aerodynamic influence on close proximity flight for both quadrotors, specifically the effect of downwash from the upper quadrotor on the lower quadrotor. 

\begin{figure}[ht!]
    \centering    \includegraphics[width=0.45\textwidth]{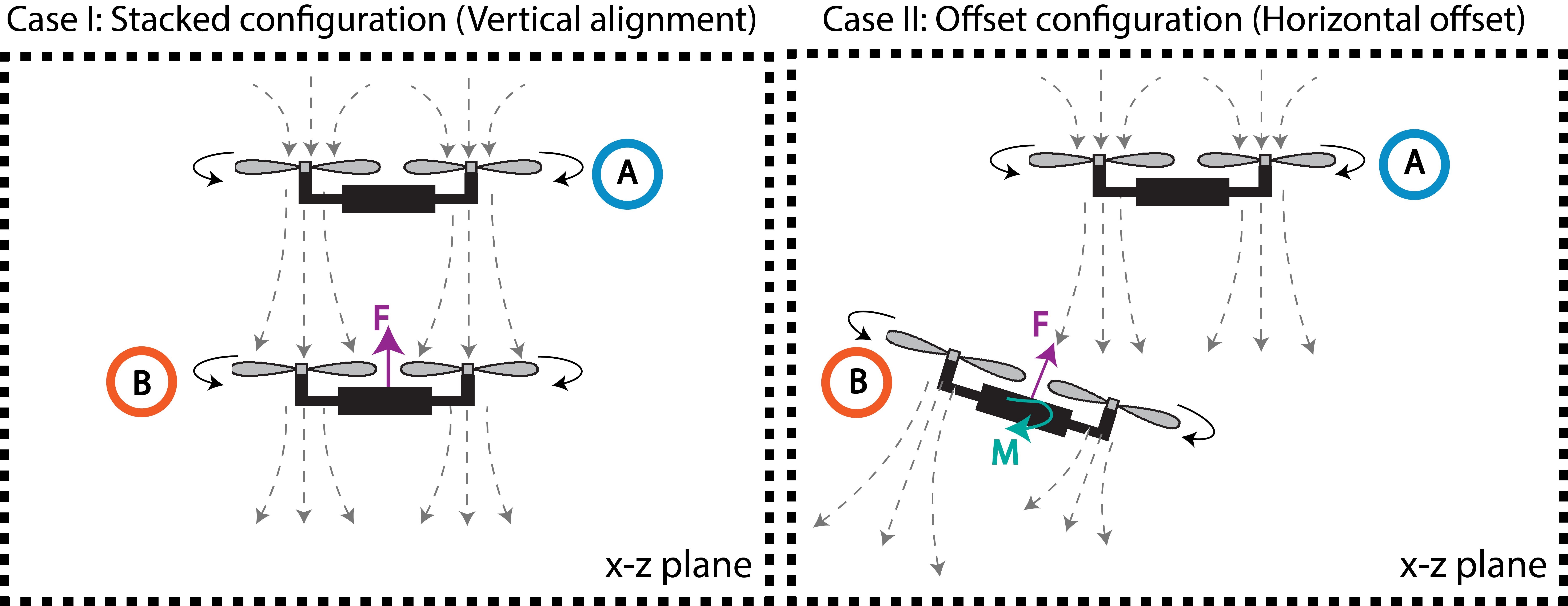}
    \caption{Close proximity quadrotor flight under the influence of downwash; Case I (left) depicts vertical alignment with quadrotors (A) and (B) in a stacked configuration. Case II (right) illustrates quadrotor (B) horizontally offset from quadrotor (A). Black arrows adjacent to the rotors indicate their direction of rotation\label{fig:case_schematic}. In this figure, $F$ represents force, while $M$ represents moment.}
\end{figure}

Two sample cases of the effects of quadrotor downwash are shown in Fig.~\ref{fig:case_schematic}. Here, two quadrotors fly one above the other in a vertically aligned configuration (left) and a horizontally offset configuration (right). 

In this work, we present an in-depth data analysis from experiments performed on quadrotors mounted to a custom translation stage that allows precise relative positioning of the quadrotors within a PIV test environment while preventing roll and yaw. 
These experiments were conducted in a wind-free environment to ensure that the setup closely mimics the free-flight conditions of the quadrotors in an indoor laboratory flight space. The study focuses on the effective change in thrust and pitch moment of the quadrotors due to the influence of downwash, as they were horizontally and vertically separated as in Fig.~\ref{fig:force_moment_setup}.

In case I, quadrotors (A) and (B) are in a stacked configuration (Fig.~\ref{fig:case_schematic}). Vertically aligned rotors in this configuration rotate in the same direction, and the jet from the upper downwash-generating quadrotor (A) directly impacts the propellers of the quadrotor below (B), significantly affecting its performance and stability. This is because the downwash jet from the upper quadrotor (A) modifies the inflow of the lower quadrotor (B) due to the substantial induced airflow generated below the downwash-generating quadrotor. Thus, the thrust, $F_z$, generated by the lower quadrotor (B) is expected to be strongly affected when the quadrotors are in a stacked configuration.

Case II considers quadrotors (A) and (B) with some horizontal offset. With increasing horizontal displacement from the initial stacked case, the pitch moment, $M_y$, acting on the lower quadrotor from the downwash-generating quadrotor above increases, peaking in magnitude at the maximum moment arm displacement. Destabilization induced by this pitching moment requires a restoring moment to retain stability. As the horizontal separation increases from this offset, the pitch moment decreases until no influence is felt by either quadrotor. 

\subsection{Experiment Methods}
\begin{figure}[h!]
    \centering
    \includegraphics[width=.47\textwidth]{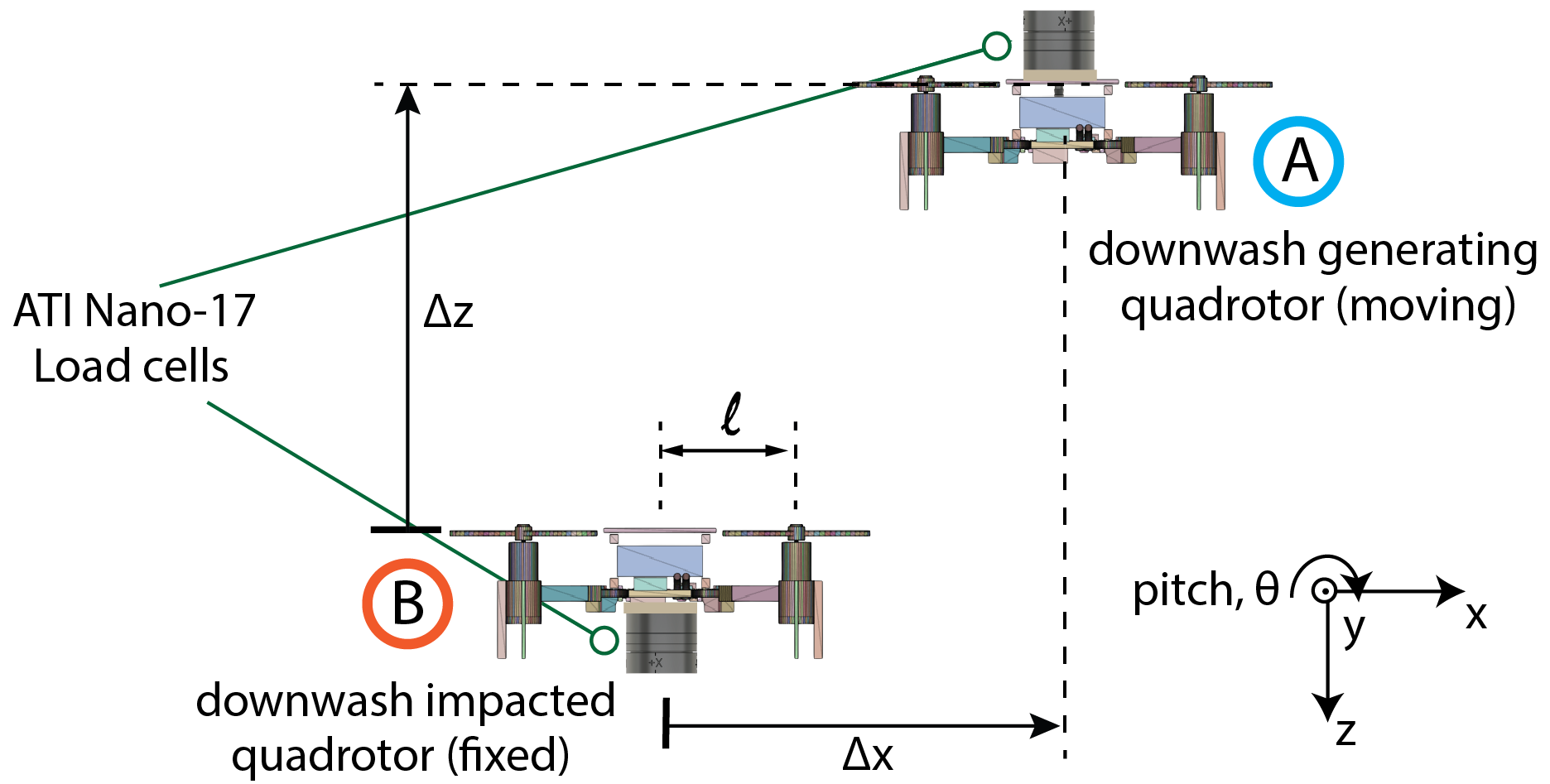}
    \caption{Coordinate system and setup schematic to measure aerodynamic interactions on forces and moments acting on a pair of quadrotors.
    \label{fig:force_moment_setup}}
\end{figure}

The experimental setup (Fig.~\ref{fig:force_moment_setup}) consists of two Crazyflie quadrotors mounted on a $x-z$ translation stage. Quadrotor (B) is fixed while quadrotor (A) traverses a wide range of horizontal ($\Delta{x}$) and vertical ($\Delta{z}$) separations using stepper motor controllers with a position accuracy of 100 $\mu$m. Both quadrotors are mounted on 6 DOF force/torque transducers (ATI Nano-17). 

All four motors on both quadrotors are operated with a constant DC voltage supply (through a tether) of 4 V at a converted bit-PWM signal for hover, commanded via Crazyradio~\cite{HonigFlightROS}. This generates a total thrust sufficient to maintain steady-state hover in free flight. At this setting, the rotors rotate at approximately 320 Hz. Note that the Crazyflie quadrotor control hardware suite maintains constant voltage, not constant rotational frequency. 
We conducted two sets of experiments, one for force and moment measurements and another for velocity measurements. The specific cases for each of these measurements are outlined below:

\subsubsection{Force \& Moment measurements\label{subsubsec:F&M_measurements}}

\begin{figure*}[t]
    \centering
    \begin{subfigure}[t]{0.24\textwidth}
        \centering
        \includegraphics[width=\linewidth]{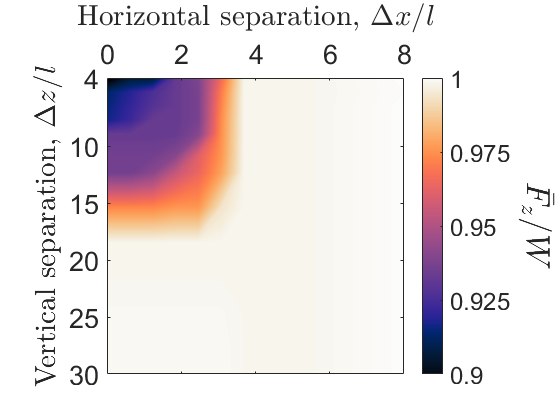}
        \caption{Thrust, $\bar{F_z}$ upper}
        \label{fig:FM_upper_A}
    \end{subfigure}
    \hfill
    \begin{subfigure}[t]{0.24\textwidth}
        \centering
        \includegraphics[width=\linewidth]{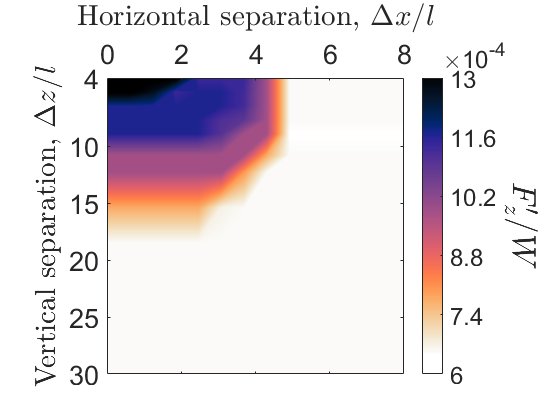}
        \caption{Thrust std. dev., $F_z'$ upper}
        \label{fig:FM_upper_B}
    \end{subfigure}
    \hfill
    \begin{subfigure}[t]{0.24\textwidth}
        \centering
        \includegraphics[width=\linewidth]{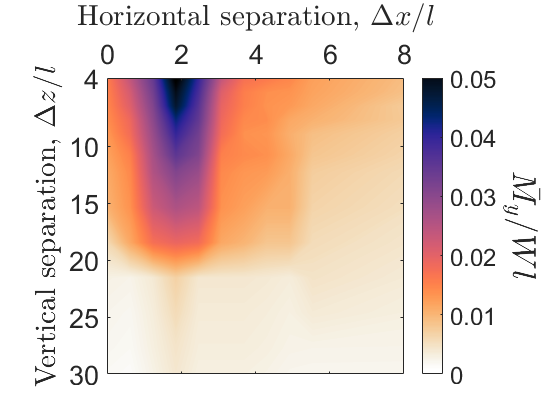}
        \caption{Moment, $\bar{M_y}$ upper}
        \label{fig:FM_upper_C}
    \end{subfigure}
    \hfill
    \begin{subfigure}[t]{0.24\textwidth}
        \centering
        \includegraphics[width=\linewidth]{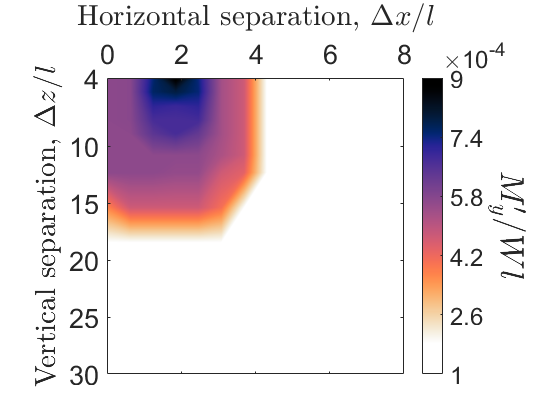}
        \caption{Moment std. dev., $M_y'$ upper}
        \label{fig:FM_upper_D}
    \end{subfigure}
    \caption{Mean and standard deviations (unsteadiness) of forces and moments experienced by \emph{upper quadrotor} generating downwash at a separation (${\Delta}x$, ${\Delta}z$) above the lower quadrotor. Horizontal and vertical separations are normalized by the arm length, $l$, in Fig.~\ref{fig:force_moment_setup}
    \label{fig:FM_upper}}
\end{figure*}

\begin{figure*}[t]
    \centering
    \begin{subfigure}[t]{0.24\textwidth}
        \centering
        \includegraphics[width=\linewidth]{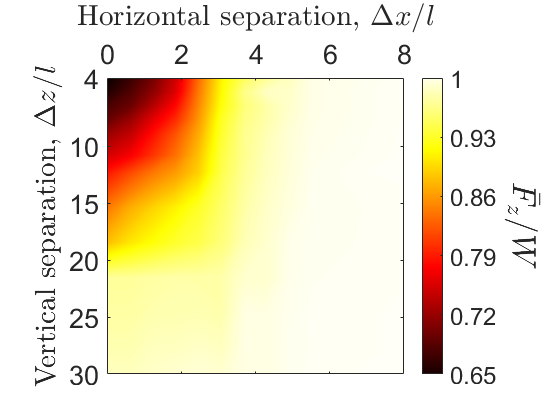}
        \caption{Thrust, $\bar{F_z}$ lower}
        \label{fig:FM_lower_A}
    \end{subfigure}
    \hfill
    \begin{subfigure}[t]{0.24\textwidth}
        \centering
        \includegraphics[width=\linewidth]{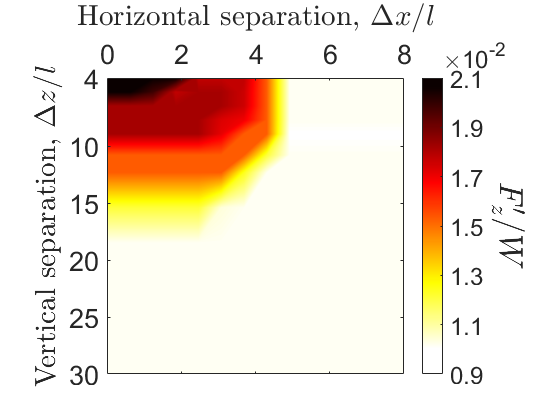}
        \caption{Thrust std. dev., $F_z'$ lower}
        \label{fig:FM_lower_B}
    \end{subfigure}
    \hfill
    \begin{subfigure}[t]{0.24\textwidth}
        \centering
        \includegraphics[width=\linewidth]{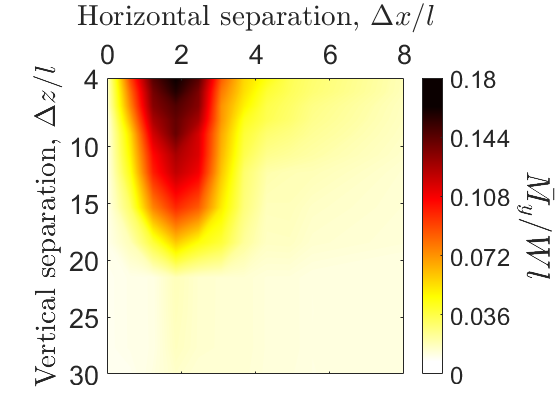}
        \caption{Moment, $\bar{M_y}$ lower}
        \label{fig:FM_lower_C}
    \end{subfigure}
    \hfill
    \begin{subfigure}[t]{0.24\textwidth}
        \centering
        \includegraphics[width=\linewidth]{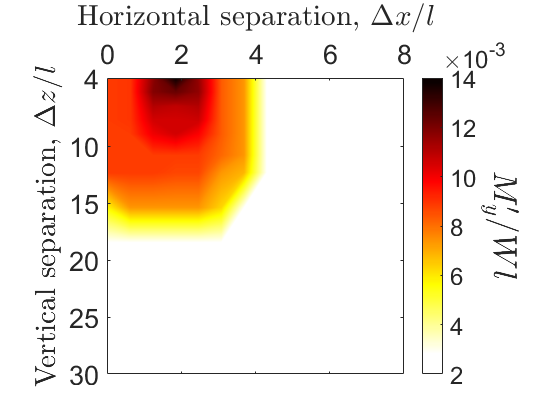}
        \caption{Moment std. dev., $M_y'$ lower}
        \label{fig:FM_lower_D}
    \end{subfigure}
    \caption{Mean and standard deviations (unsteadiness) of forces and moments experienced by \emph{lower quadrotor} due to downwash of quadrotor positioned above at a separation (${\Delta}x$, ${\Delta}z$). Horizontal and vertical separations are normalized by the arm length, $l$, in Fig.~\ref{fig:force_moment_setup}.
    \label{fig:FM_lower}}
    \vspace{-0.1in}
\end{figure*}

The relative position between quadrotors (${\Delta}x$, ${\Delta}z$) was varied over a rectangular test matrix ranging 0.1 m $\leq {\Delta}x \leq$ 0.6 m horizontally, and 0.1 m $\leq {\Delta}z \leq$ 1.1 m vertically. Evaluating a larger range for vertical versus horizontal separation was guided by the axis-aligned ellipsoid model ~\cite{HonigDownwashTraj} and the cylindrical model ~\cite{NathanGraspTestbed}, which suggest that downwash develops more significantly in the axial (vertical) direction, and is further supported by experimental data in our study (cf. Figs.~\ref{fig:FM_upper}, \ref{fig:FM_lower}, \ref{fig:single_cf_piv_composite_A}). 
We recorded force and moment measurements from both quadrotors at a sampling rate of 20 kHz for 30 seconds at each relative position; each case was repeated three times, and the standard deviation was computed from these trials. 

\subsubsection{Velocity measurements\label{subsubsec:V_measurements}}  
Extreme instances of the downwash effect, illustrated in Fig.~\ref{fig:case_schematic}, guided our selection of cases for velocity measurements. Velocity measurements were performed in the stacked configuration (case I) and the offset configuration (case II) at the closest vertical separation (${\Delta}z/l = 4$) as in force and moment measurements. To complement these, we also included a farther vertical separation case (${\Delta}z/l = 12.5$) for both the stacked and offset configurations. This resulted in a total of four multi-vehicle PIV configurations for velocity measurements. Additionally, a fifth case was included to capture the flow field below a single quadrotor, providing insights into the merging of individual rotor jets and their wake structure in the absence of vehicle-vehicle interaction, serving as a baseline for comparison with the multi-vehicle cases. 

\begin{figure}[h!]
    \centering
    \includegraphics[width=.47\textwidth]{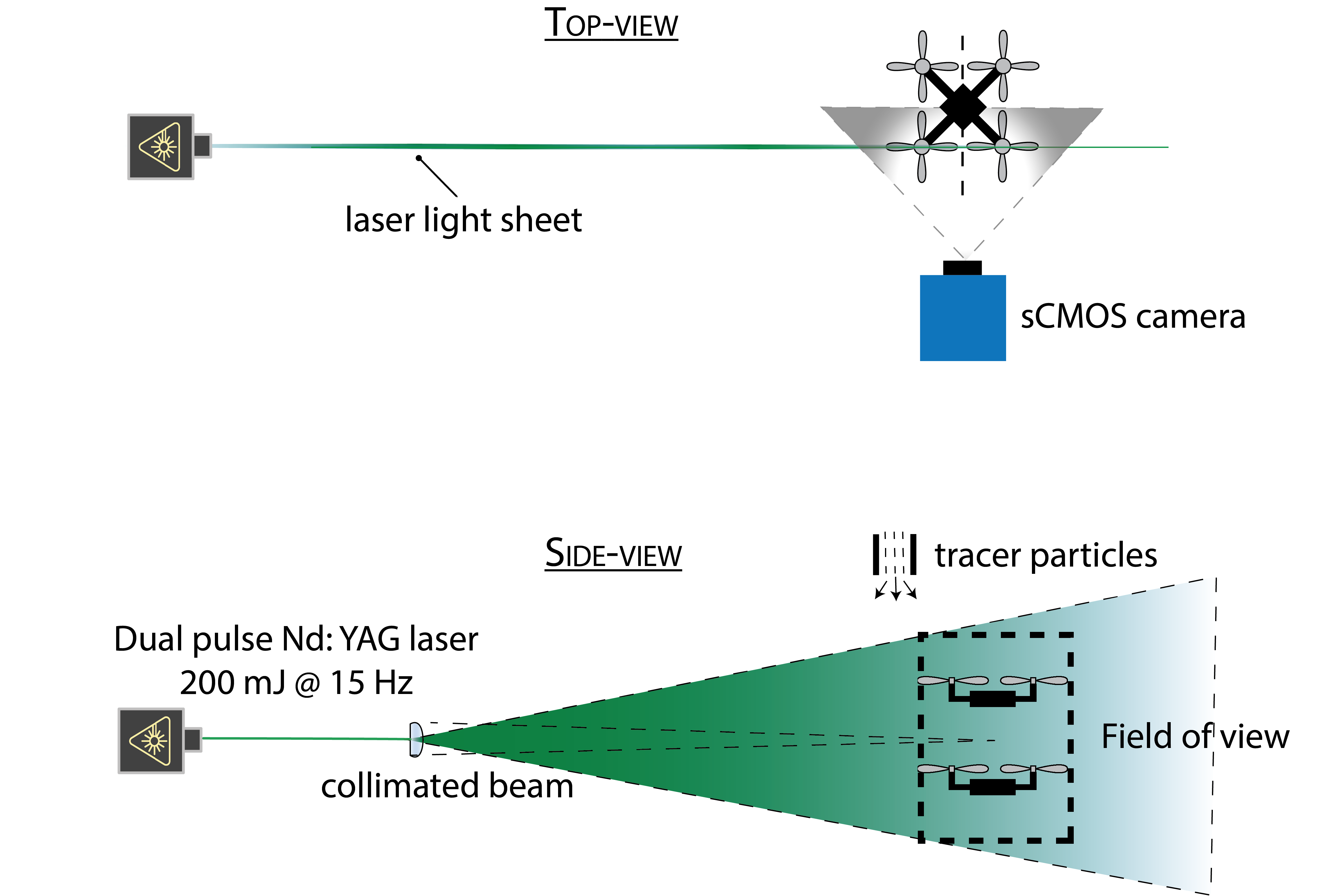}
    \caption{PIV setup for quadrotor downwash velocity data acquisition. The laser sheet illuminates the tracer particles along the plane (side-view), slicing through the centers of the front two rotors of the quadrotors (top-view), capturing the flow field along that plane. 
    \label{fig:PIV_setup}}
\end{figure}

PIV provides a quantitative measure of velocities generated above and below the rotors. Small ($\sim \mu$m) tracer droplets of Di-Ethyl-Hexyl-Sebacic-Acid-Ester (DEHS) were suspended in the flow field and tracked under the influence of downwash. The PIV system used a Quantel Evergreen dual-cavity Nd:Yag laser (532 nm, max pulse energy: 200 mJ @ 15 Hz) to illuminate the tracer particles. The laser sheet was aligned to illuminate a plane passing through the center of the front two rotors in the cross configuration, as shown by the top view in Fig.~\ref{fig:PIV_setup}. Reflective surfaces, including motors, were properly masked and coated with anti-reflective paint to ensure detailed flow visualization. Particle motion was captured by a LaVision Imager sCMOS camera (2560 $\times$ 2160 pixels) equipped with a Nikon 35 mm lens, offering a field of view of 400 $\times$ 300 mm ($x \times z$). A total of 1000 image pairs containing velocity fields were recorded, representing 70 seconds of data, and processed using DaVis v10 (LaVision), using multipass correlations with a final interrogation area of 32 $\times$ 32 pixels and a 75\% overlap.

To cover the far-field wake structure for a single quadrotor, and regions above and below both quadrotors in stacked and offset configurations, the quadrotors were raised and lowered on the $x-z$ translational stage, keeping their relative position, the laser light sheet, and the camera fixed, to image the flow fields in sections. The average velocity field from each section was combined to form a composite flow field covering a larger vertical range. Each section had a minimum overlap with adjacent sections of 100 mm, one-third of the $z$-direction measurement range, to ensure a smooth transition and maintain spatial resolution when stitching sections together. 

\section{Results and Discussion}

\subsection{Forces \& Moments\label{subsubsec:F&M_results}}

In Figs.~\ref{fig:FM_lower_A} and~\ref{fig:FM_lower_C}, the forces and moments experienced by the lower quadrotor are dramatically altered by the downwash generated from the upper vehicle. The thrust generated by the quadrotors is normalized by their weight, $W$ ($\approx$ 0.265 N), and the pitch moment, $M_y$ is normalized by $W l$, where $l$ is the horizontal distance from the Crazyflie quadrotor's body center to a motor (32.5 mm) depicted in Fig.~\ref{fig:force_moment_setup}. The effective downwash intensity peaks for the lower quadrotor when both vehicles are placed in the stacked configuration (vertical alignment), and at the closest vertical separation (${\Delta}z/l = 4$), the thrust significantly drops to 65\% of its nominal hover thrust. As the downwash-inducing upper quadrotor translates away horizontally and vertically relative to the lower quadrotor, it recovers its nominal hover thrust, balancing its lift with its weight, $W$. The lower quadrotor's thrust attains unaltered levels at (${\Delta}x/l \approx 3$) and (${\Delta}z/l \approx 19$). Regions within these limits represent areas of downwash influence, consistent with the axially-aligned ellipsoid model~\cite{HonigDownwashTraj}, which states that the downwash effect is more pronounced in the axial direction than in the lateral direction.

For ${{\Delta}z}/l > 19$, there is a negligible effect from downwash on the bottom quadrotor. At this vertical separation, the thrust generated by the lower quadrotor ranges between 95\% - 98\% of its hover thrust, as seen in Fig.~\ref{fig:FM_lower_A}. For ${{\Delta}x}/l > 3.7$, the effect of downwash on thrust has nearly diminished. In contrast, we observe that the downwash-producing upper quadrotor experiences only a slight reduction in its thrust as seen in Fig.~\ref{fig:FM_upper_A}. Even at the closest vertical separation (${\Delta}z/l = 4$) from the lower quadrotor, the upper quadrotor retains 90\% of its original hover thrust. 

In addition to the magnitudes of normalized thrust due to downwash, we quantify the unsteadiness associated with downwash based on the standard deviation of the normalized thrust, $F_{z}'$. The unsteadiness associated with downwash lingers until vertical separation ${{\Delta}z}/l \approx 17$ and horizontal separation ${{\Delta}x}/l \approx 4.5$ (cf. Figs.~\ref{fig:FM_upper_B} and \ref{fig:FM_lower_B}). Similar to the effect on thrust, the unsteadiness remains pronounced when the quadrotors align vertically up to (${{\Delta}z}/l \approx 14$). The standard deviation of relative thrust, $F_{z}'$ is of much higher magnitude for the lower quadrotor compared to the upper quadrotor, the upper quadrotor experiences turbulence approximately two orders of magnitude smaller than the lower quadrotor. The flow eventually becomes steady, as seen by the diminishing influence of fluctuating forces for $14 < {{\Delta}z}/l < 17$. The vertical separations contain regions of unsteadiness spread over a range that is three times that of separations on the horizontal scale. For separations ${{\Delta}z}/l > 17$ and ${{\Delta}x}/l > 4.5$, the unsteadiness associated with downwash can be considered negligible.

The pitch moment magnitudes, $\bar{M}_y$, experienced by the upper and lower quadrotors as functions of vehicle separations (${\Delta}x$, ${\Delta}z$) are shown in Figs.~\ref{fig:FM_upper_C} and \ref{fig:FM_lower_C}. Both quadrotors begin to experience a pitch moment when their centerlines are horizontally displaced. This effect grows until they reach a horizontal separation of ${{\Delta}x}/l = 2$. At this specific configuration, the upper quadrotor's downwash is partially offset, with its left rotors aligned with the lower quadrotor's right rotors, as shown in the offset configuration (case II) in Fig.~\ref{fig:case_schematic}. This specific configuration results in the largest moment arm, maximizing the pitch moment based on the definition of moments as the product of force and its perpendicular distance from the pivot location. 
However, the non-interacting rotors operate in less a disturbed airflow, partially preserving vehicle thrust, resulting in lesser overall thrust reduction as seen in Fig.~\ref{fig:FM_lower_A} for this configuration, compared to the stacked configuration in Fig.~\ref{fig:case_schematic} (case I), where downwash from the upper quadrotor is concentrated directly on all four rotors of the lower quadrotor uniformly.
The upper quadrotor experiences a minimal change in pitch moment---only one-third of that observed by the lower quadrotor, as seen in Fig.~\ref{fig:FM_upper_C}. 

As seen in Figs.~\ref{fig:FM_upper_D} and \ref{fig:FM_lower_D}, the standard deviation of the pitch moment, $M'_y$, is non-zero, even with no horizontal displacement. 
 
The unsteady moments peak at ${{\Delta}x}/l$ = 2 before decaying again, becoming negligible by ${{\Delta}x}/l \approx 4$. For vertical separations ${{\Delta}z}/l > 16$, the flow becomes established, leading to a steady presence. Similar to force fluctuations, it can be observed that the moment fluctuations are about two orders of magnitude smaller for the upper quadrotor compared to the lower quadrotor (cf.~Figs.~\ref{fig:FM_upper_D} and \ref{fig:FM_lower_D}). This can be attributed to the intensity of downwash and the turbulent fluctuations associated with it, which primarily act on the lower quadrotor. 

\subsection{Velocity fields}

1000 frames of the two-dimensional velocity field ($u$, $v$) were collected using PIV for each configuration. The lateral component along the $x$-axis is denoted as $v$, while the axial velocity component along the $z$-axis is denoted as $u$, based on the orientation shown in Fig.~\ref{fig:force_moment_setup}. Note that $u$ is positive, pointing toward the ground. The resultant velocity fields are averaged over 1000 frames - $(\bar{u}, \bar{v})$ - to analyze the spatial structure of the flow fields, including jet merging and spreading patterns. 

For single-vehicle flow fields, both axial and lateral velocity components are presented for a comprehensive understanding of flow dynamics. However, for the multi-vehicle cases, we focus on the downwash speed, $\bar{U}$ ($\bar{U} = \sqrt{\bar{u}^2 + \bar{v}^2}$), which captures the contributions from both axial and lateral components.

Velocities are normalized by the induced velocity, $U_i$, which estimates downward airflow generated by the rotor blades of a propeller as they produce lift according to the actuator disk theory~\cite{LeishmanHeli}. While $U_i$ provides a simplified estimate of the flow directly beneath the propeller and is assumed to be uniform immediately below the rotor in actuator disk theory, it does not necessarily represent the actual flow velocity, as the downwash velocity downstream, $U$, evolves as a function of both $x$ and $z$. Assuming quasi-steady flow based on actuator disk theory~\cite{LeishmanHeli}, at the rotor disk, the vehicle thrust, $F_z$, relates to the induced velocity, $U_{i}$: 
\begin{equation}
U_{i}=\sqrt{\frac{F_z}{2{\rho}An}} ,
\label{eqn:v_ind}
\end{equation}
where $\rho$ is the air density, $n$ represents the number of propellers on the aerial vehicle, and $A = \pi r^2$ is the area swept by a propeller, where $r$ is the radius of the propeller.

\begin{figure*}[t]
    \centering
    \begin{subfigure}[t]{0.20\textwidth}
        \centering
        \includegraphics[width=\linewidth]{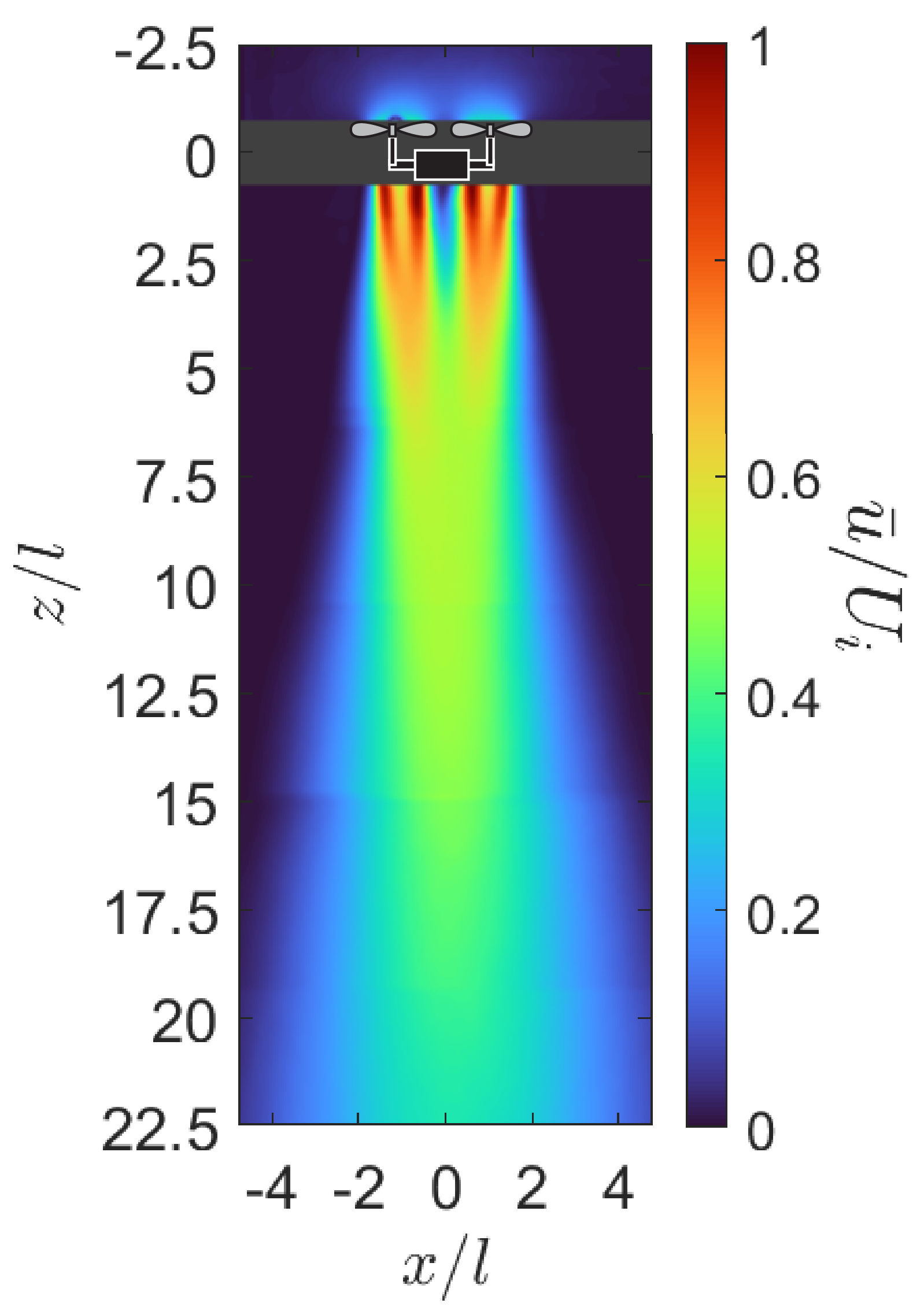}
        \caption{Axial velocity ($\bar{u}$) field}
        \label{fig:single_cf_piv_composite_A}
    \end{subfigure}
    \hfill
    \begin{subfigure}[t]{0.29\textwidth}
        \centering
        \includegraphics[width=\linewidth]{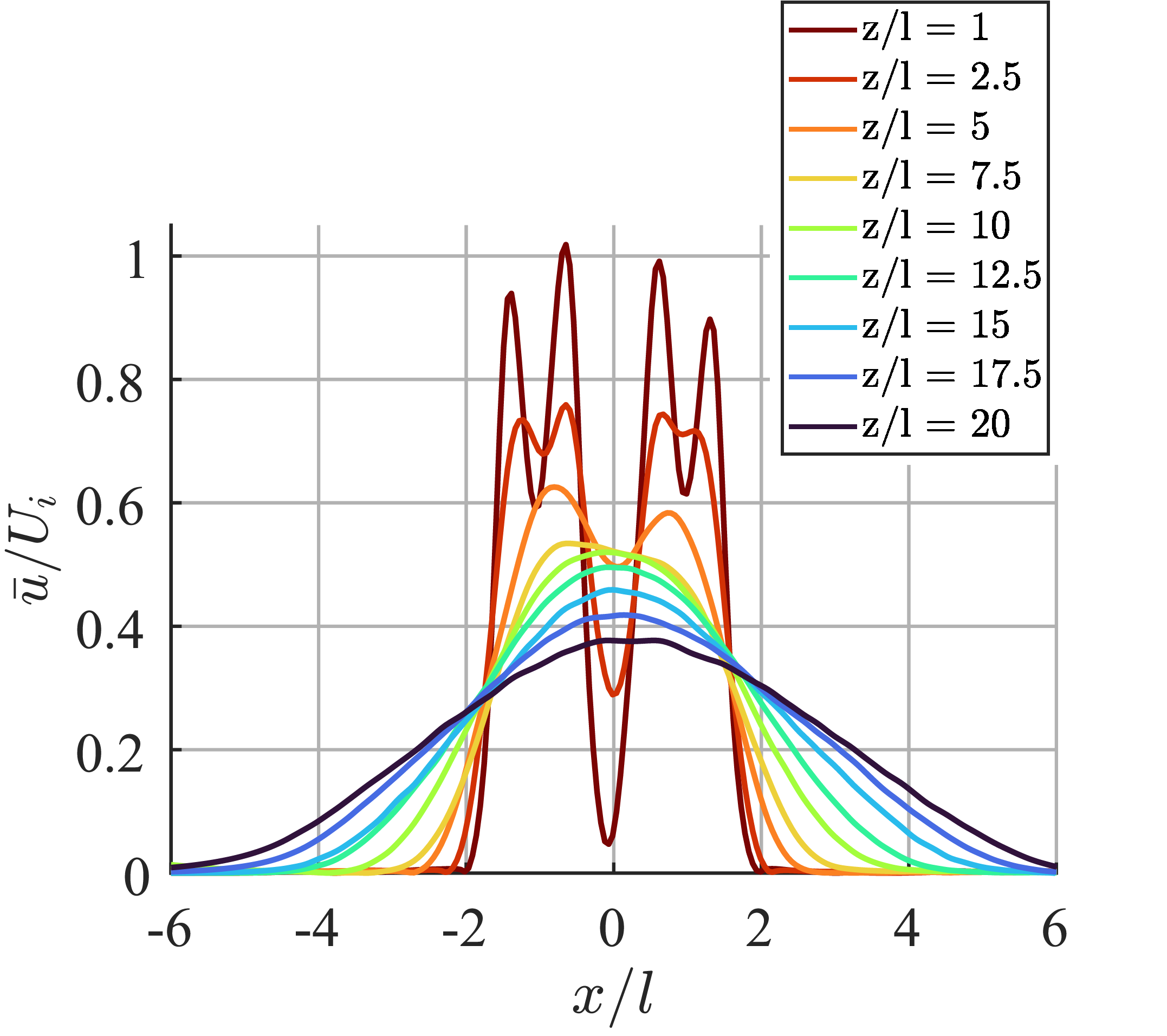}
        \caption{Axial velocity ($\bar{u}$) profiles}
        \label{fig:single_cf_piv_composite_B}
    \end{subfigure}
    \hfill
    \begin{subfigure}[t]{0.20\textwidth}
        \centering
        \includegraphics[width=\linewidth]{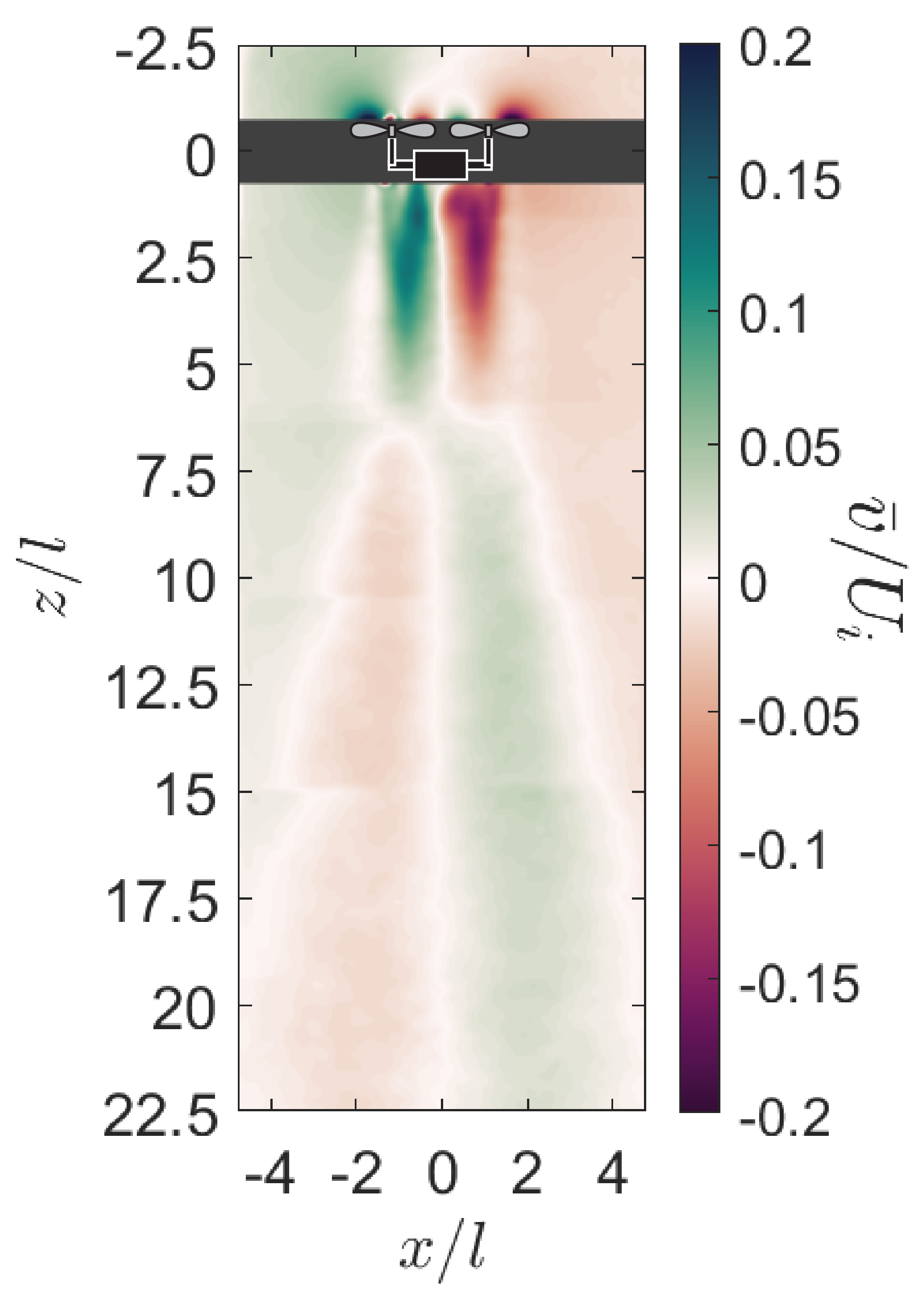}
        \caption{Lateral velocity ($\bar{v}$) field}
        \label{fig:single_cf_piv_composite_C}
    \end{subfigure}
    \hfill
    \begin{subfigure}[t]{0.29\textwidth}
        \centering
        \includegraphics[width=\linewidth]{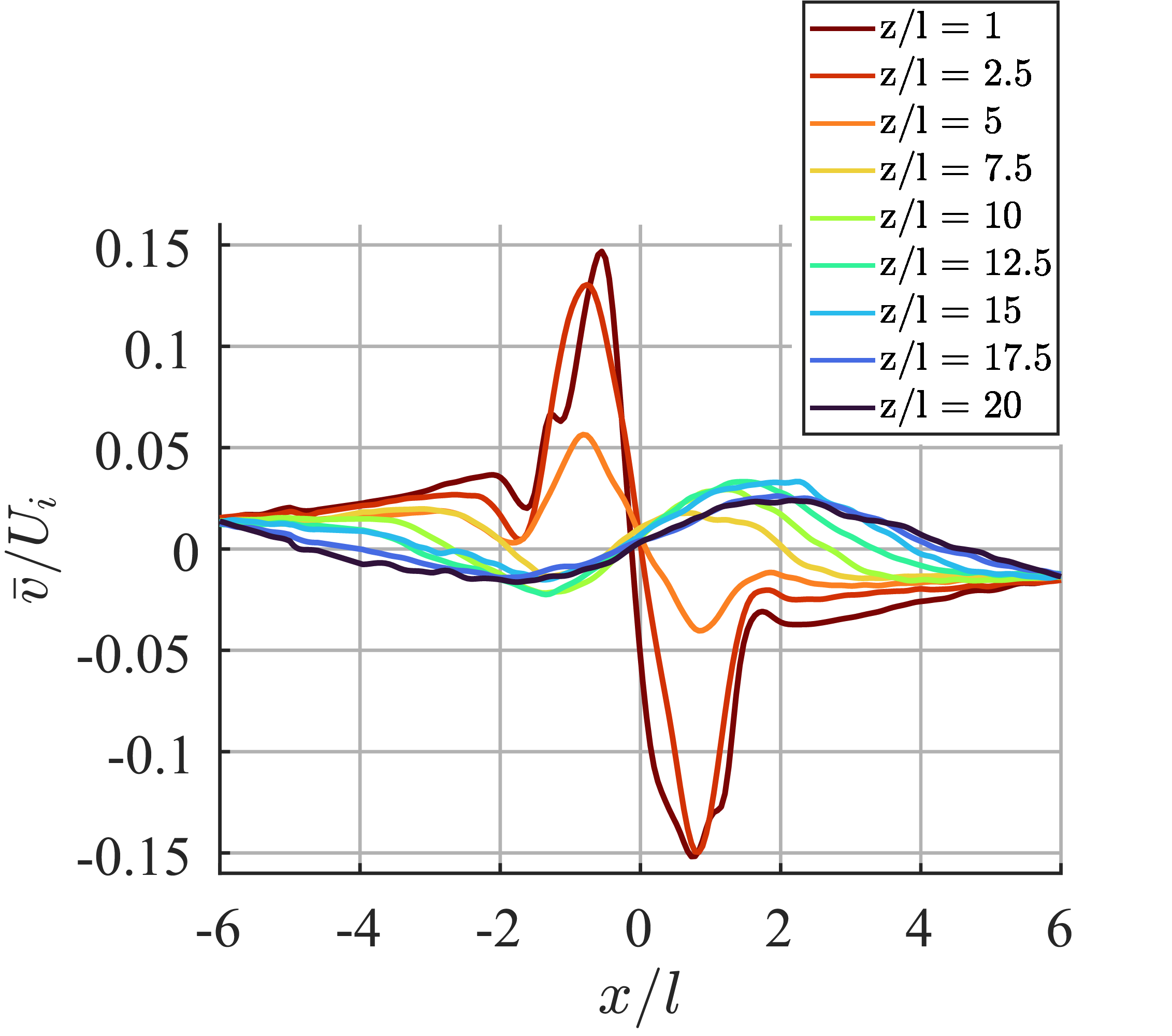}
        \caption{Lateral velocity ($\bar{v}$) profiles}
        \label{fig:single_cf_piv_composite_D}
    \end{subfigure}
    \caption{Time-averaged axial ($\bar{u}$) and lateral downwash ($\bar{v}$) velocities, normalized by the induced velocity ($U_i$) for Crazyflie quadrotor in hover (Eq.~\eqref{eqn:v_ind}), along with corresponding velocity profiles at various downstream ($z/l$) locations. Flow field images include overlaid illustrations of Crazyflie quadrotors for reference.
    \label{fig:single_cf_PIV}}
    \vspace{-0.1in}
\end{figure*}

\subsubsection{Single vehicle near-field flow characteristics}

First, we present the downwash velocity data from a single quadrotor. 
Multiple time-averaged image sequences were stitched together to form a composite flow field extending over several rotor diameters below the vehicle, allowing us to characterize both near-field and far-field flow features with a detailed spatial resolution of 1.6 mm (0.05$l$). The flows that we observed are broadly similar to computational fluid dynamics (CFD) computations of multirotor wakes~\cite{DiazComputationalAero, YoonComputationalAerodynamicModeling}. The measurements show similarities to the PIV of time-averaged flow fields of Carter et al.~\cite{CarterInfluenceGroundCeiling} and those of flow-probe measurements from Bauersfeld et al.~\cite{BauersfeldRoboticsMeetsFluidDynamics}.

The propeller's rotation produces the inflow velocity above the rotors, generating a suction effect that draws in the surrounding air. The inflow velocity is roughly one-fifth of the induced velocity below the rotors and can be noticed above the left and right propellers of the quadrotor in Fig.~\ref{fig:single_cf_piv_composite_A}. Similar flow structures have been quantified for modeling the ceiling effect~\cite{CarterInfluenceGroundCeiling} and near-field flow analyses~\cite{KiranDownwashSeparation, WolfeQuadShakethebox}.

Rotor jets are visible as a high-speed annulus below the rotor blades and a low-speed core below the rotor hub~\cite{DekkerGroundEffect, KiranDownwashSeparation}. The downwash speed and fluctuations in the near-field have been previously studied~\cite{KiranDownwashSeparation}. However, here we present an analysis of the entire flow field. A dead zone exists between the rotor jets in the near-field immediately below the quadrotor body, as in Fig.~\ref{fig:single_cf_piv_composite_A}, due to the spacing between the quadrotor propellers and the quadrotor body's presence. This is consistent with CFD simulations~\cite{DiazComputationalAero, YoonComputationalAerodynamicModeling} and experiments~\cite{WolfeQuadShakethebox, BauersfeldRoboticsMeetsFluidDynamics}. Figure~\ref{fig:single_cf_piv_composite_B} shows the individual rotor jets merging into a single jet as we move axially downstream in the centerline region ($x/l = 0$). The transition of individual rotor jets into a single turbulent jet observed in this study, $z/l = 6.5$ ($\approx$ 2.3 motor-to-motor distances), is consistent with Bauersfeld et al., which reports a similar transition occurring after 2.5 motor-distance length scales~\cite{BauersfeldRoboticsMeetsFluidDynamics}. This agreement strengthens the understanding that quadrotor downwash develops into a turbulent jet past the near-field region. Beyond this merge point, $z_m/l$ (i.e., $z/l = 6.5$), the maximum downwash speed exists at the centerline region, expanding downstream further along the axial direction into the far-field ($z/l>6.5$) region. 

The time-averaged lateral velocity ($\bar{v}$) in Fig.~\ref{fig:single_cf_piv_composite_C} remains zero at the centerline in both the near- and far-field regions. Immediately below the rotors, the flow moves inward towards the vehicle axis, but the polarity of the lateral velocity switches beyond the merge point and the merged jet starts to spread outwards, its intensity weakening beyond the axial merge point.

Figures~\ref{fig:single_cf_piv_composite_B} and~\ref{fig:single_cf_piv_composite_D} show the axial and lateral velocity profiles, respectively, by taking slices along the PIV measurements of time-averaged downwash flow fields from Figs.~\ref{fig:single_cf_piv_composite_A} and~\ref{fig:single_cf_piv_composite_C}, respectively, at several values of $z/l$. Axial velocity profiles maintain symmetry about the centerline ($x/l = 0$), representing two independent rotor wakes that merge to form a unified jet. The axial velocity increases rapidly at the centerline and reaches its maximum magnitude as the rotor jets coalesce. At the merge point, the profiles exhibit a decay in magnitude while maintaining a self-similar shape and broadening laterally further downstream into the far-field.

The velocity profile transitions from a highly non-uniform flow near the rotors, where multiple localized peaks appear due to individual rotor wakes. In the near field (up to $z/l \approx 5$), these localized maxima gradually merge into two dominant peaks (bimodal distribution) by $5\leq z/l \leq 6.5$. Eventually, for $z/l \ge 6.5$, the velocity profile transitions into a shape dominated by a single peak at the centerline, with a broader lateral spread and gradually tapering towards the edges.

\subsubsection{Single vehicle far field flow characteristics - Comparison with turbulent jet theory}

\begin{figure}[tb]
    \centering    \includegraphics[width=0.3\textwidth]{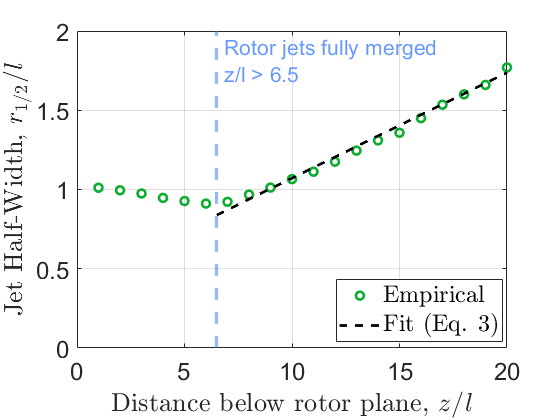}
    \caption{Downwash jet half-width, $r_{\frac{1}{2}}(z)$, plotted against downstream locations below the rotor plane, $z/l$. Note that once jets merge (right of dashed light blue line at $z/l > 6.5$), the merged jet half-width grows linearly with $z$ downstream.
\label{fig:jet_width_growth}}
\end{figure}

\begin{figure}[tb]
    \centering    \includegraphics[width=0.3\textwidth]{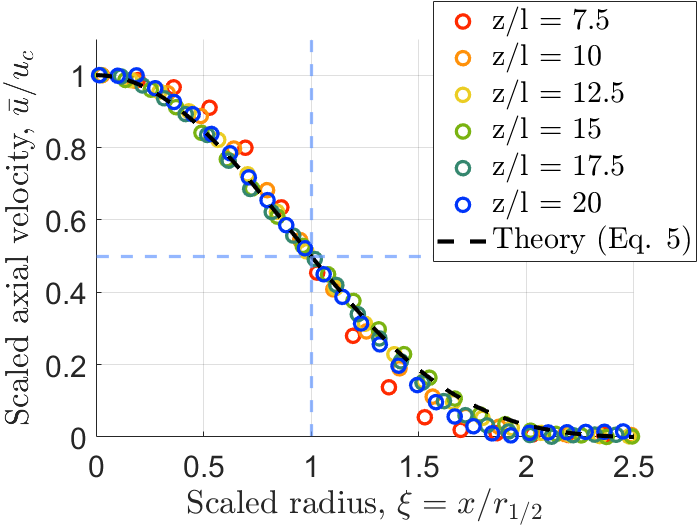}
    \caption{Scaled axial velocities plotted against the lateral position, normalized by the corresponding jet half-widths. Empirical data from multiple far-field axial locations, where the flow has transitioned into a turbulent jet, aligns well with the theoretical prediction in Eq.~\eqref{eqn:axial_similarity}.
    \label{fig:axial_self_sim}}
\end{figure}

Our experimental far-field flow measurements are compared with classical turbulent jet theory \cite{PopeTurbulentFlows}, which models flow emanating from a concentrated momentum source. In this regime, the jet half-width ($r_{\frac{1}{2}}$) quantifies the lateral expansion of the downwash across the jet's cross-section~\cite{PopeTurbulentFlows}. This parameter is defined as the lateral distance from the centerline ($x$) where the axial velocity is one-half of its centerline maximum (Fig.~\ref{fig:cover_image} (left)):
\begin{equation}
    \bar{u}\left(r_{\frac{1}{2}}(z),z\right)\:=\:\frac{1}{2}u_c\left(z\right) \ .
    \label{eqn:jet_half-width}
\end{equation}
Notably, this self-similar framework is valid only once the flow has transitioned to the far-field region ($z/l > 6.5$), where it behaves as a turbulent jet.
The turbulent jet theory establishes that at any specific downstream location, the lateral position corresponding to the jet half-width represents precisely where the axial velocity decreases to half of its centerline maximum value. This relationship is illustrated by the dashed light blue lines in Fig.~\ref{fig:axial_self_sim}. Beyond the merge point, the velocity profiles transition to a single-peaked distribution as seen for $z/l>6.5$ in Fig.~\ref{fig:single_cf_piv_composite_B}, and the jet half-width increases linearly with $z$ at a constant rate, $S$, as shown in Fig.~\ref{fig:jet_width_growth} with the relationship:
\begin{equation}
    r_{\frac{1}{2}}(z)=S(z-z_0),
\label{eqn:jet_width_growth}
\end{equation}
where $z_0$ represents the virtual origin determined by tracing the flow back to where the spread theoretically begins, i.e., a reference point located above the quadrotor. The black dashed line in Fig.~\ref{fig:jet_width_growth} represents the function form fit for jet half-width growth. 

\begin{figure}[tb]
    \centering    \includegraphics[width=0.3\textwidth]{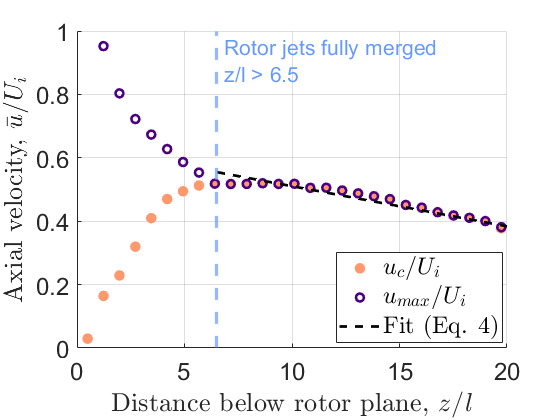}
    \caption{Axial velocity, $\bar{u}$ in a merged turbulent jet decays linearly with $z$. Plotted in peach colors are the centerline velocities ($u_c$) at the axial centerline along each downstream slice below the rotor plane ($z/l$). The purple symbols show maximum axial velocity ($u_{max}$) along each downstream location. Note that once jets merge ($z/l > 6.5$), both $u_{max}$ and $u_{c}$ overlap and they collectively decay linearly with $z$, consistent with turbulent jets.
 \label{fig:axial_decay}}
\end{figure} 

To elucidate jet merging, we consider the axial centerline (i.e., at  $x/l = 0$) velocity, $u_{c}$, and the maximum axial velocity, $u_{max}$, along each downstream slice, $z/l$. The centerline velocity, $u_{c}$ starts near zero in the dead zone just below the rotors and gradually gains magnitude, peaking at half the induced velocity at $z/l = 6.5$, which can be observed in Figs.~\ref{fig:single_cf_piv_composite_B} and \ref{fig:axial_decay}, then drops linearly farther downstream. 
The maximum axial velocity, $u_{max}$, attains its maximum magnitude
immediately below the rotor plane and stays steady briefly at $z/l = 6.5$. 
At the merge point ($z_m$), we expect $u_{max}$ and $u_{c}$ to coincide, which indicates that the individual rotor jets have merged, with the maximum axial velocity at the axial centerline further downstream of this unified jet. Figure~\ref{fig:axial_decay} shows where rotor jet merging results in an overlap of $u_{c}$ and $u_{max}$ starting at $z/l = 6.5$, beyond which $u_{c}$ and $u_{max}$ are identical in value and coincide decaying linearly with $z$, corroborating our prediction that the individual rotor jets have coalesced. This is influenced by the velocity profiles attaining a single peak for $z/l > 6.5$ with radial tapering, the magnitude of which decays downstream ($z/l$), while spreading over a wider lateral extent as in Fig.~\ref{fig:single_cf_piv_composite_B}. By definition, the maximum velocity of a turbulent jet happens at the centerline location \cite{PopeTurbulentFlows}, consistent with our findings (Fig.~\ref{fig:axial_decay}).

The black dashed line in Fig.~\ref{fig:axial_decay} represents the function form fit for axial decay of centerline velocity, $u_c$, for a turbulent jet~\cite{PopeTurbulentFlows}:
\begin{equation}
    u_c\left(z\right)=u_0\frac{Bd}{z-z_0},
    \label{eqn:axial_decay}
\end{equation}
where $u_0$ corresponds to the initial jet velocity (set to 1 as in \cite{BauersfeldRoboticsMeetsFluidDynamics}), $B$ represents an empirical constant tied to the decay rate, $d$ stands for the exit diameter of the jet, and $z_0$ indicates the virtual origin defined earlier. Both scaling analyses for the normalized jet half-width growth \eqref{eqn:jet_width_growth} and normalized centerline velocity decay \eqref{eqn:axial_decay} are consistent with related work~\cite{BauersfeldRoboticsMeetsFluidDynamics}; parameters for each scaling are compared in Table~\ref{table:fit_params}. The higher resolution (0.05$l$) in our measurement allows for a detailed analysis of the flow field, supported by a velocity uncertainty of 0.6 mm/s, derived from a particle displacement error of 0.2 px per interrogation window at a sampling rate of 15 Hz. 

\begin{table}[tb]
\centering
 \begin{tabular}{| c | c | c | c | c ||} 
 \hline
 Fit parameters & $Bd$ & $S$ & $z_0$ \\ [0.5ex] 
 \hline\hline
 Present study & 9.508 & 0.0667 & -6.0585 \\ 
 \hline
 Bauersfeld et al. \cite{BauersfeldRoboticsMeetsFluidDynamics} & 10.11 &  0.07668 & -5.817\\ [1ex] 
 \hline
 \end{tabular}
 \caption{\textbf{Single vehicle far-field fit parameters} \label{table:fit_params}}
\end{table}

\begin{figure}[tb]
    \centering    \includegraphics[width=0.3\textwidth]{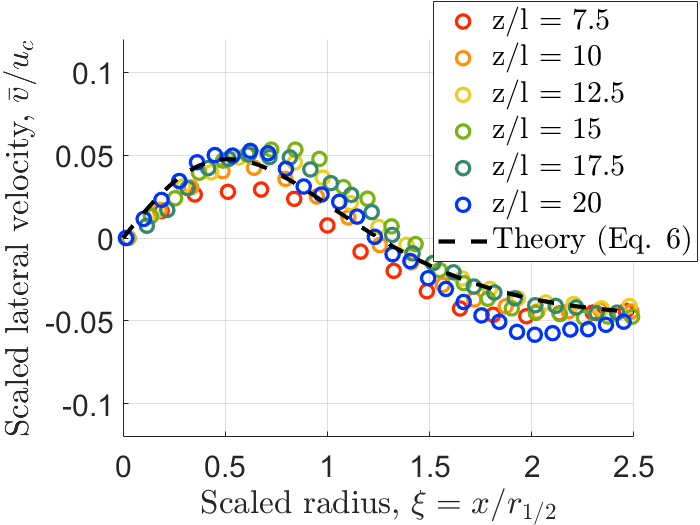}
    \caption{Scaled lateral velocities plotted against the lateral position, normalized by the corresponding jet half-widths. Empirical data from multiple far-field axial locations, where the flow has transitioned into a turbulent jet, aligns well with the theoretical prediction in Eq.~\eqref{eqn:lateral_similarity}.
    \label{fig:lateral_self_sim}}
\end{figure}

\begin{figure*}[tbh]
    \centering
    \begin{subfigure}[t]{0.20\textwidth}
        \centering
        \includegraphics[width=\linewidth]{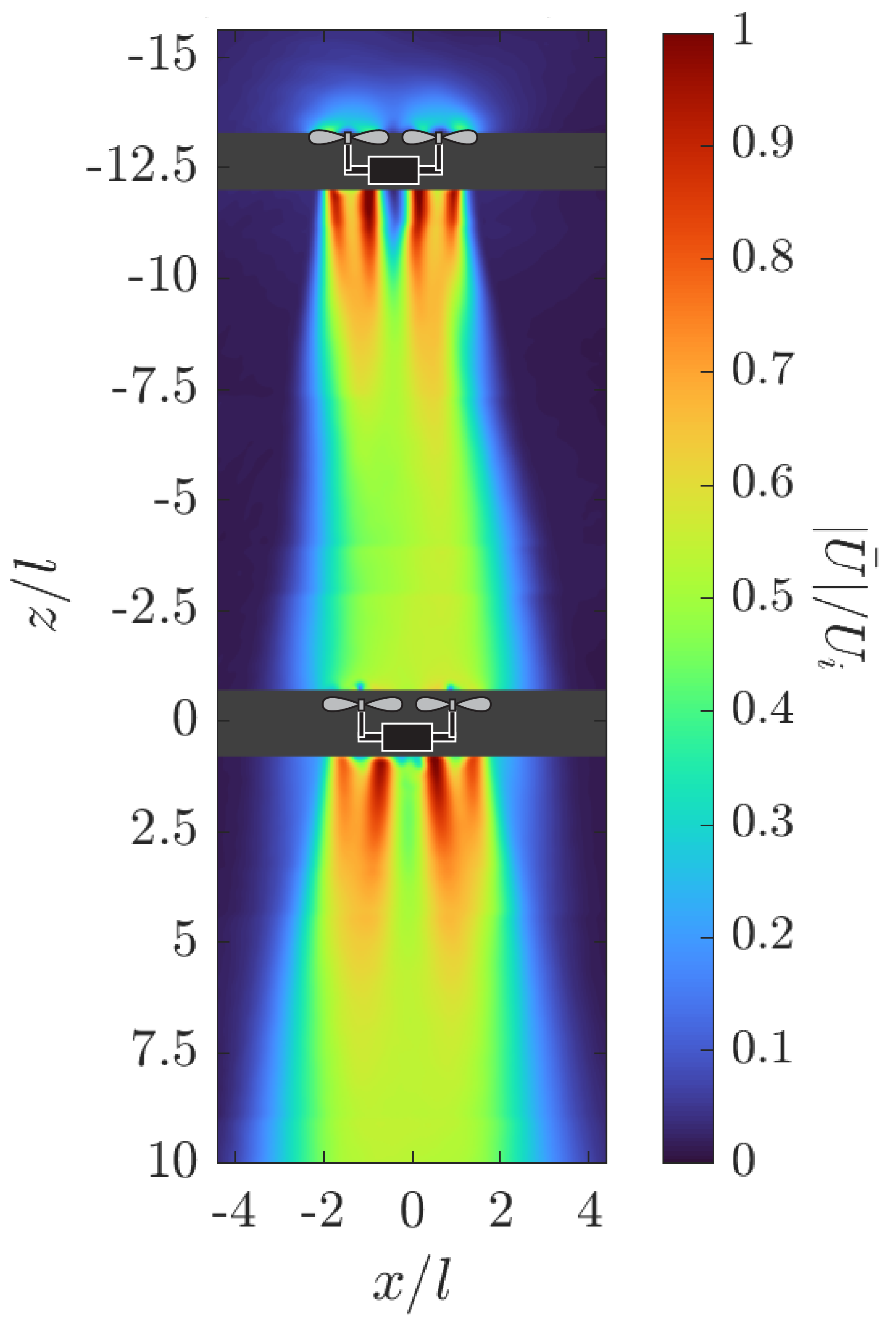}
        \caption{Flow field, ${\Delta}z/l = 12.5$}
        \label{fig:multi_cf_maxF_downwash_speed_A}
    \end{subfigure}
    \hfill
    \begin{subfigure}[t]{0.24\textwidth}
        \centering
        \includegraphics[width=\linewidth]{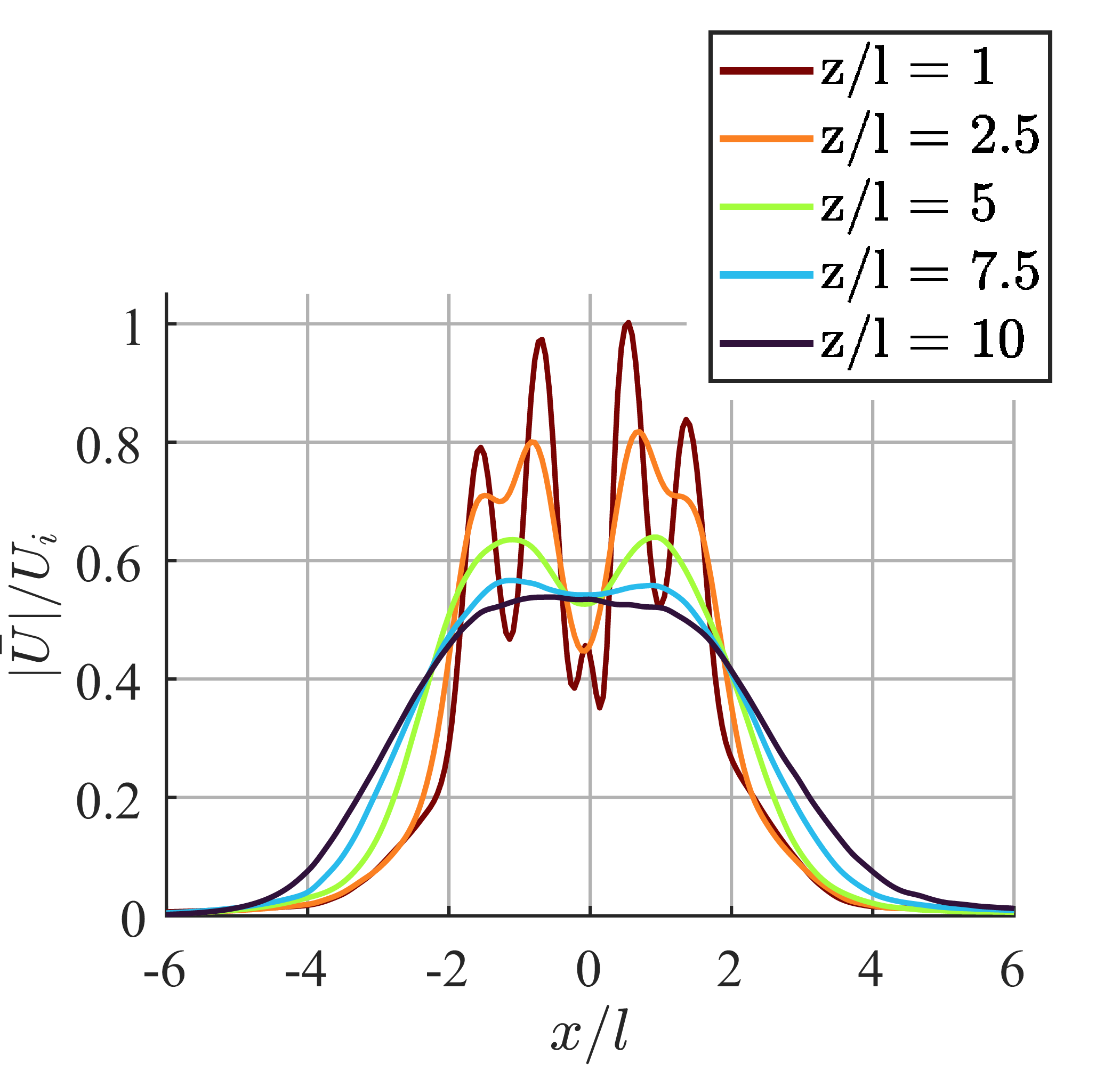}
        \caption{Velocity profile, ${\Delta}z/l = 12.5$}
        \label{fig:multi_cf_maxF_downwash_speed_B}
    \end{subfigure}
    \hfill
    \begin{subfigure}[t]{0.20\textwidth}
        \centering
        \includegraphics[width=\linewidth]{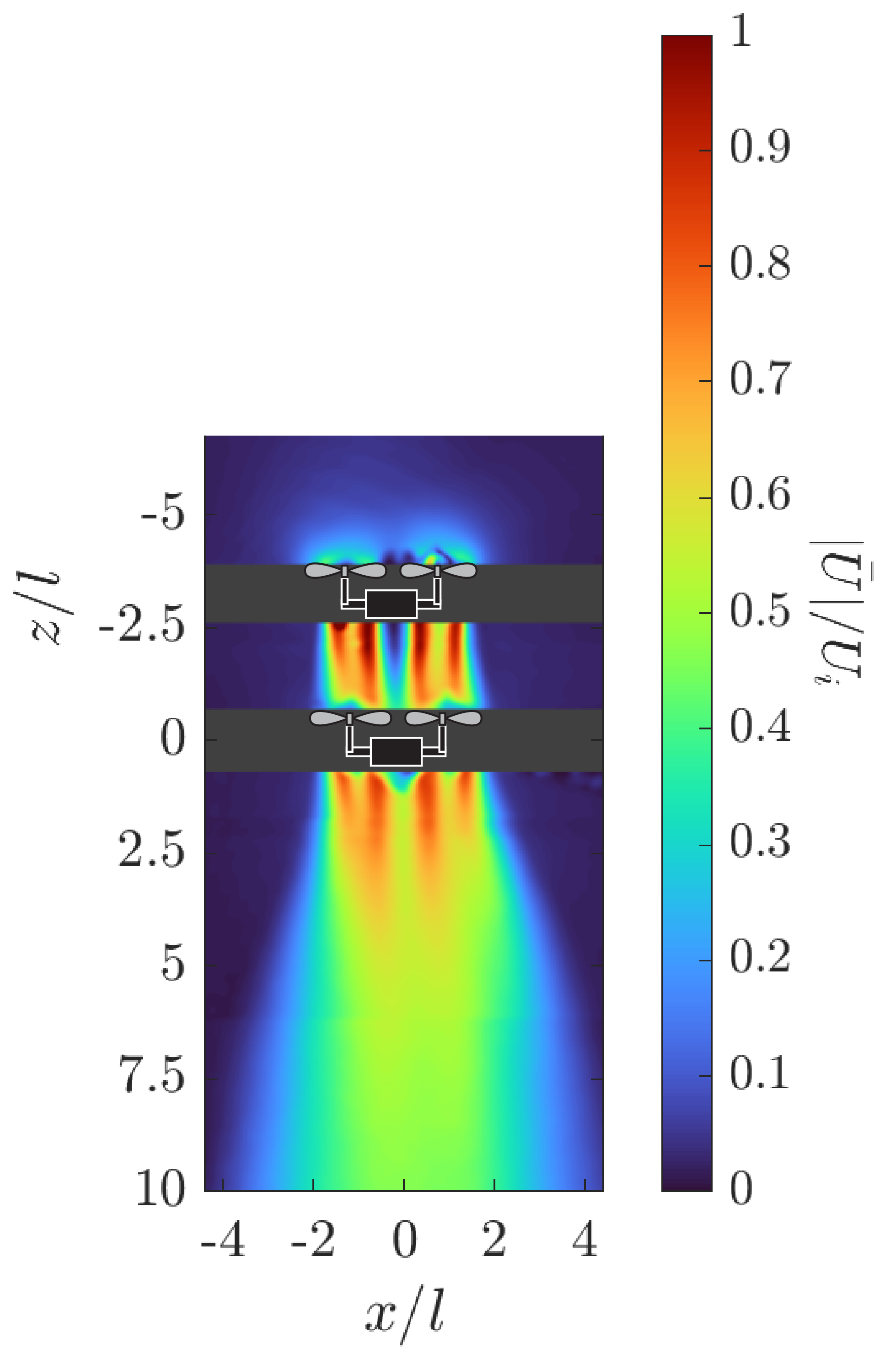}
        \caption{Flow field, ${\Delta}z/l = 4$}
        \label{fig:multi_cf_maxF_downwash_speed_C}
    \end{subfigure}
    \hfill
    \begin{subfigure}[t]{0.24\textwidth}
        \centering
        \includegraphics[width=\linewidth]{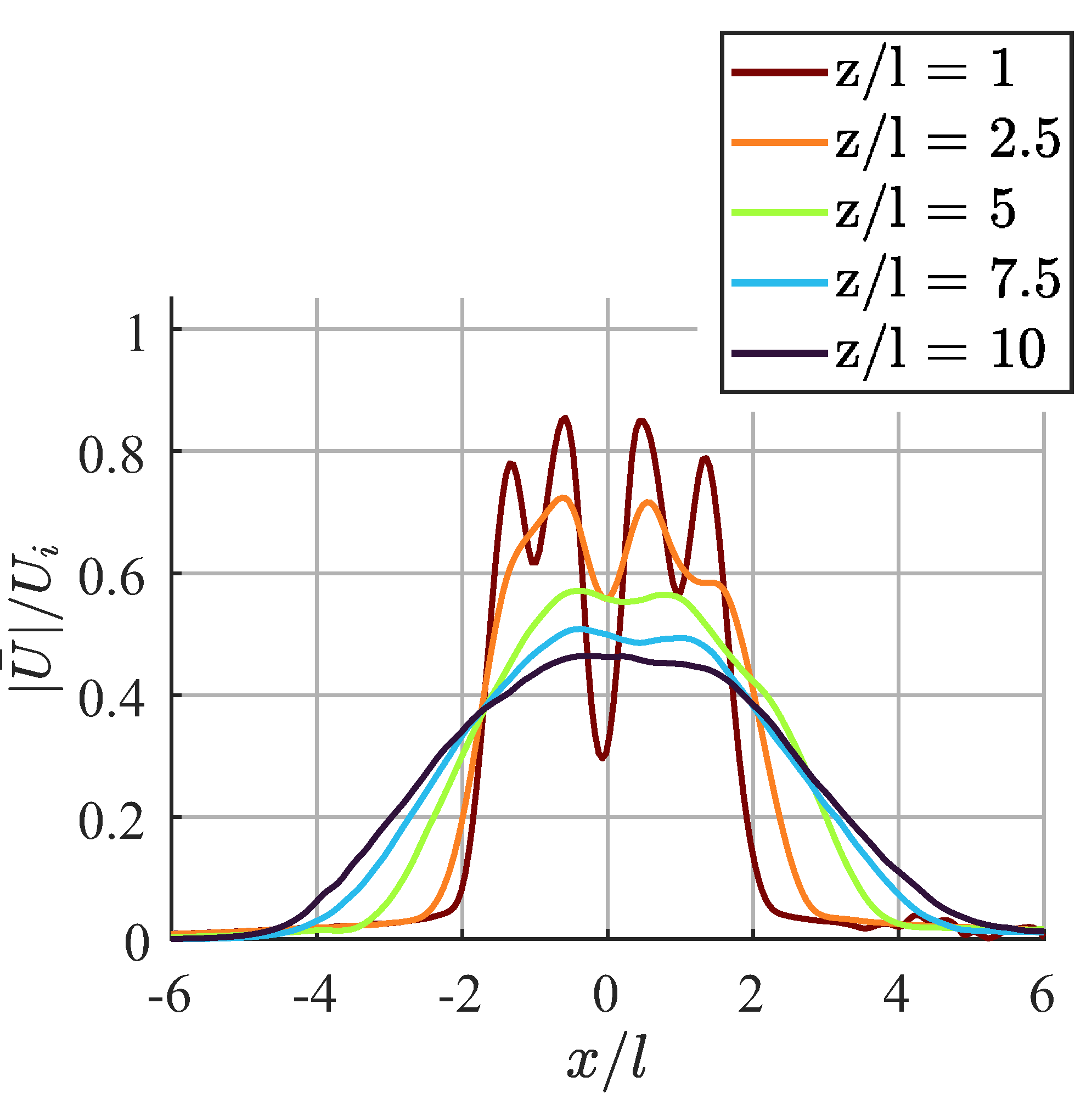}
        \caption{Velocity profile, ${\Delta}z/l = 4$}
        \label{fig:multi_cf_maxF_downwash_speed_D}
    \end{subfigure}
    \caption{Downwash speed ($|\bar{U}|$) of quadrotors in the \emph{stacked configuration} (${\Delta}x/l = 0$), along with corresponding velocity profiles below the \emph{lower quadrotor}, normalized by the induced velocity ($U_i$), at various downstream locations ($z/l$), for two vertical separations.}
    \label{fig:multi_cf_maxF}
    \vspace{-0.2in}
\end{figure*}

PIV measurements of the far-field quadrotor wake post-merge ($z/l > 6.5$) indicate that the velocity profiles attain self-similarity, i.e., the shape of the flow profiles remain consistent once appropriately scaled (cf. Figs.~\ref{fig:axial_self_sim} and \ref{fig:lateral_self_sim}), thus theory can predict its behavior accurately suggesting strong agreement 
for scaled axial \eqref{eqn:axial_similarity} and scaled lateral \eqref{eqn:lateral_similarity} velocities downstream of the merged jet in the self-similar far-field region:
\begin{equation}
    \bar{u}\left(\xi \right)\:=\frac{u_c\left(z\right)}{\left(1+\left(\sqrt{2}-1\right)\xi ^2\right)^2}\:,
    \label{eqn:axial_similarity}
\end{equation}
\begin{equation}
    \bar{v}\left(\xi \right)\:=\frac{\frac{1}{2}u_c\left(z\right)\left(\xi \:\:-\:\xi \:^3\right)}{\left(1+\xi ^2\right)^2}\:,
    \label{eqn:lateral_similarity}
\end{equation}
where 
$\xi$ represents the lateral coordinate of the downwash jet spread normalized by corresponding jet half-widths at each axial location. 
Lateral velocities exhibit relatively lower magnitudes ($\bar{v}\:\approx\:\frac{1}{20}\bar{u}$). Meanwhile, axial flow is primarily driven by the thrust generated by the rotors, resulting in a higher magnitude, directed jet downstream as observed in Fig.~\ref{fig:single_cf_piv_composite_A}.

In the far field, where self-similarity emerges, the jet half-width and velocity follow universal scaling laws, and the flow structure is dominated by turbulence. Mathematically, this can be expressed by the Reynolds number, $Re$, in terms of parameters that describe the evolving characteristics of the turbulent jet specified earlier:
\begin{equation}
    Re = \frac{u_c r_{1/2}}{\nu},
\label{eqn:reynolds_equation}
\end{equation}
where the kinematic viscosity of air, $\nu \approx  1.5 \times 10^{-5} \, \text{m}^2/\text{s}$.

At the axial location $z/l = 6.5$, where the turbulent jet achieves full merger, the centerline velocity decreases to approximately half the induced velocity, with jet half-width $r_{1/2} \approx l$. The Reynolds number at this location is calcuated as:
\begin{equation}
    Re = \frac{(4.25 \, \text{m/s}) \times (0.0325 \, \text{m})}{1.5 \times 10^{-5} \, \text{m}^2/\text{s}} \approx 9000
\label{eqn:reynolds_number}
\end{equation}
While classical turbulent jet scaling traditionally applies to high Reynolds number flows ($Re \sim 10^4$–$10^5$), where inertial forces dominate viscosity, our findings demonstrate that self-similarity persists even at lower Reynolds numbers ($Re \approx 9000$) in multirotor-generated downwash. This indicates that turbulent mixing mechanisms remain dominant despite enhanced viscous influences---a critical distinction for quadrotor aerodynamics, which typically operate in this intermediate Reynolds number regime.

The demonstrated applicability of turbulent jet scaling principles at these lower Reynolds numbers provides a robust theoretical framework for analyzing quadrotor wake interactions, developing accurate downwash models, and predicting aerodynamic behavior during formation flight in multi-vehicle systems.

\subsubsection{Multi-vehicle flow field characteristics}
Dense quadrotor flight necessitates analysis of the flow field interaction between quadrotors in close proximity. To address this, we 
analyze four configurations of multi-vehicle interactions as specified in Section~\ref{subsubsec:V_measurements}.

Figures~\ref{fig:multi_cf_maxF_downwash_speed_A} and \ref{fig:multi_cf_maxF_downwash_speed_C} shows the flow field for two vertical separations: a large vertical separation ${\Delta}z$ = 40.6 cm (${\Delta}z/l = 12.5$), and a small vertical separation ${\Delta}z$ = 13 cm (${\Delta}z/l =  4$), respectively, in the stacked configuration shown in Fig.~\ref{fig:case_schematic} case I. These flow field measurements are extracted as axial slices shown in Figs.~\ref{fig:multi_cf_maxF_downwash_speed_B} and ~\ref{fig:multi_cf_maxF_downwash_speed_D} respectively, providing detailed velocity profiles downstream of the lower quadrotor subjected to downwash effects. 

\begin{figure*}[tbh]
    \centering
    \begin{subfigure}[t]{0.20\textwidth}
        \centering
        \includegraphics[width=\linewidth]{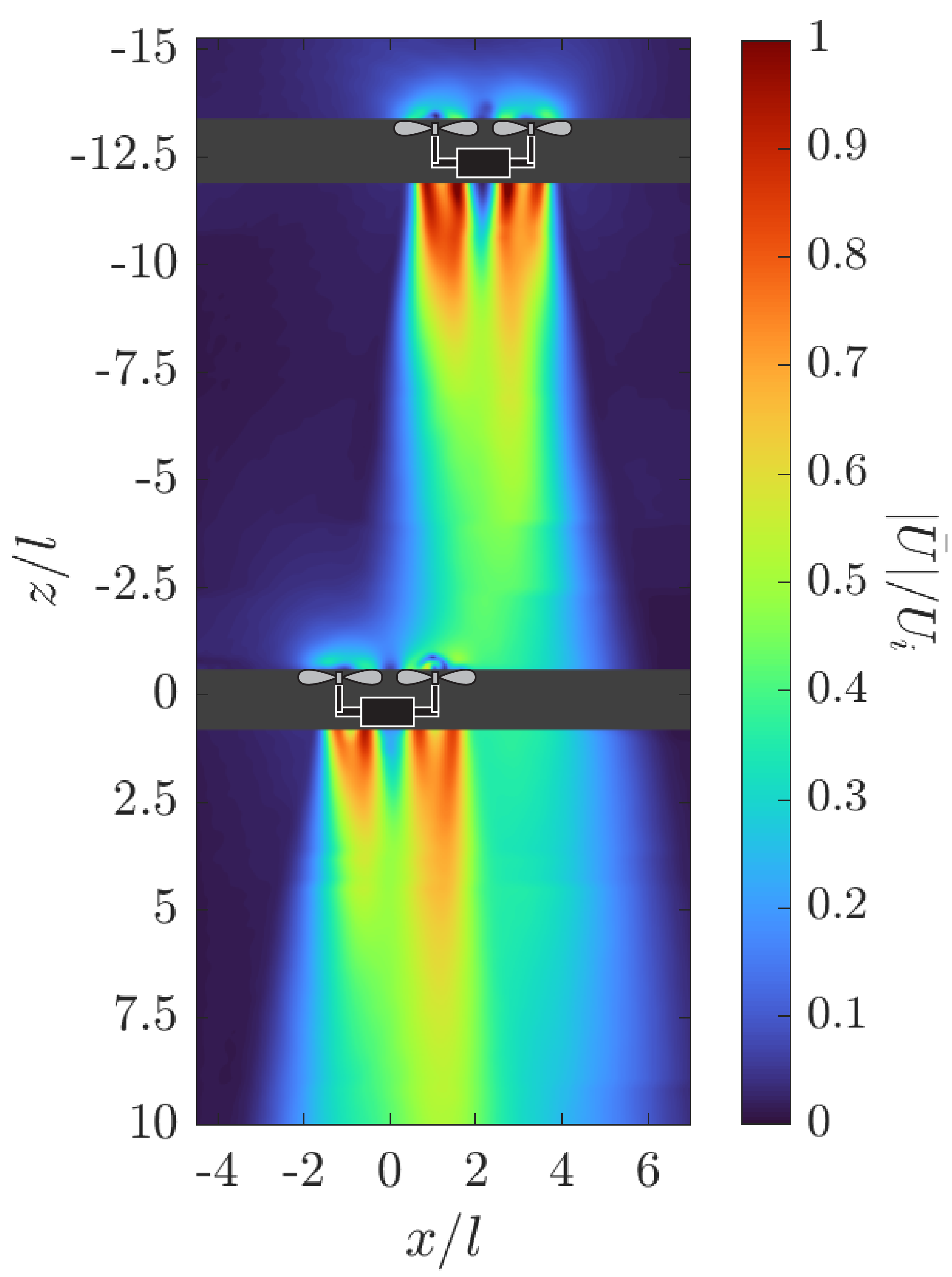}
        \caption{Flow field, ${\Delta}z/l = 12.5$}
        \label{fig:multi_cf_maxM_downwash_speed_A}
    \end{subfigure}
    \hfill
    \begin{subfigure}[t]{0.24\textwidth}
        \centering
        \includegraphics[width=\linewidth]{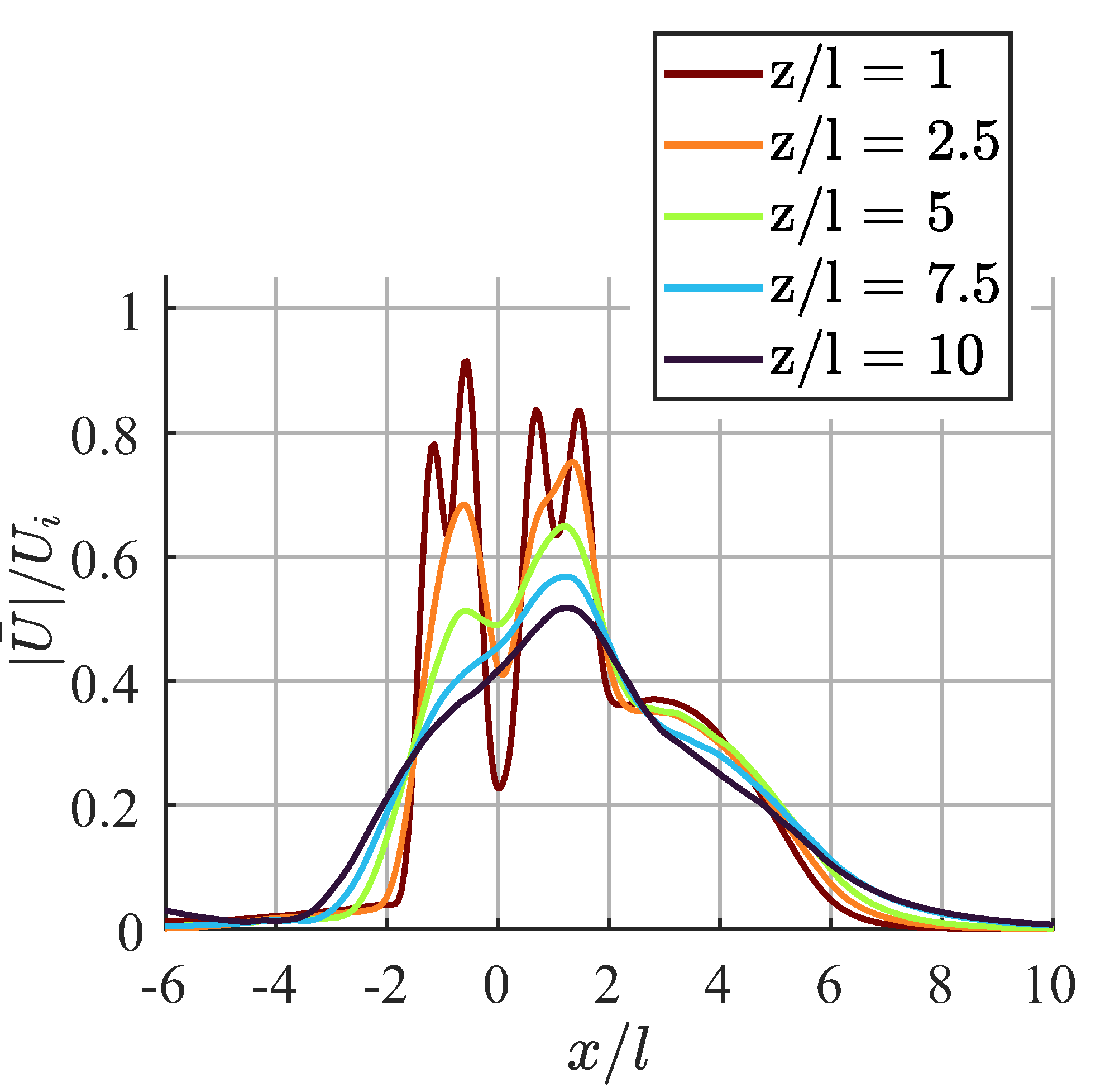}
        \caption{Velocity profile, ${\Delta}z/l = 12.5$}
        \label{fig:multi_cf_maxM_downwash_speed_B}
    \end{subfigure}
    \hfill
    \begin{subfigure}[t]{0.20\textwidth}
        \centering
        \includegraphics[width=\linewidth]{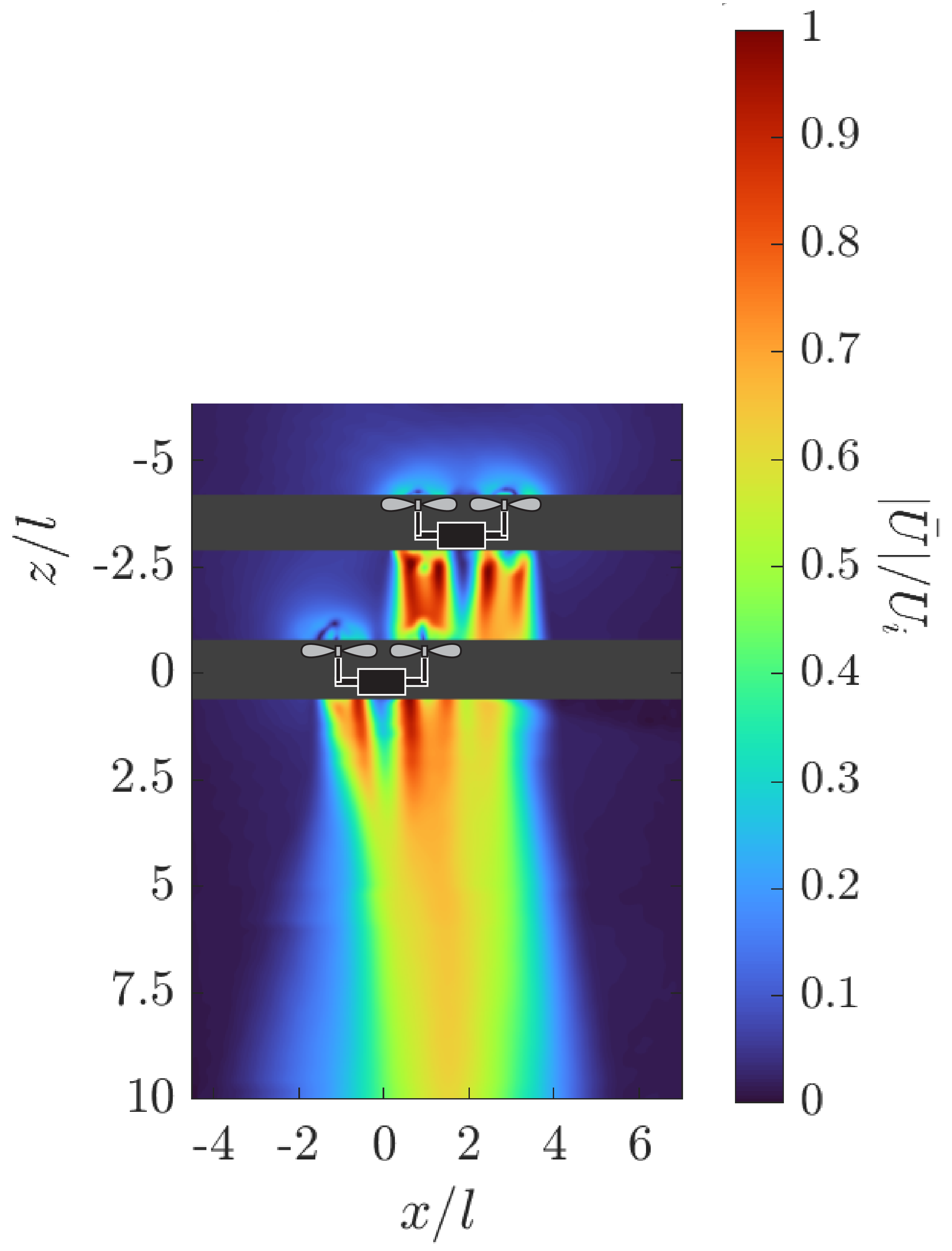}
        \caption{Flow field, ${\Delta}z/l = 4$}
        \label{fig:multi_cf_maxM_downwash_speed_C}
    \end{subfigure}
    \hfill
    \begin{subfigure}[t]{0.24\textwidth}
        \centering
        \includegraphics[width=\linewidth]{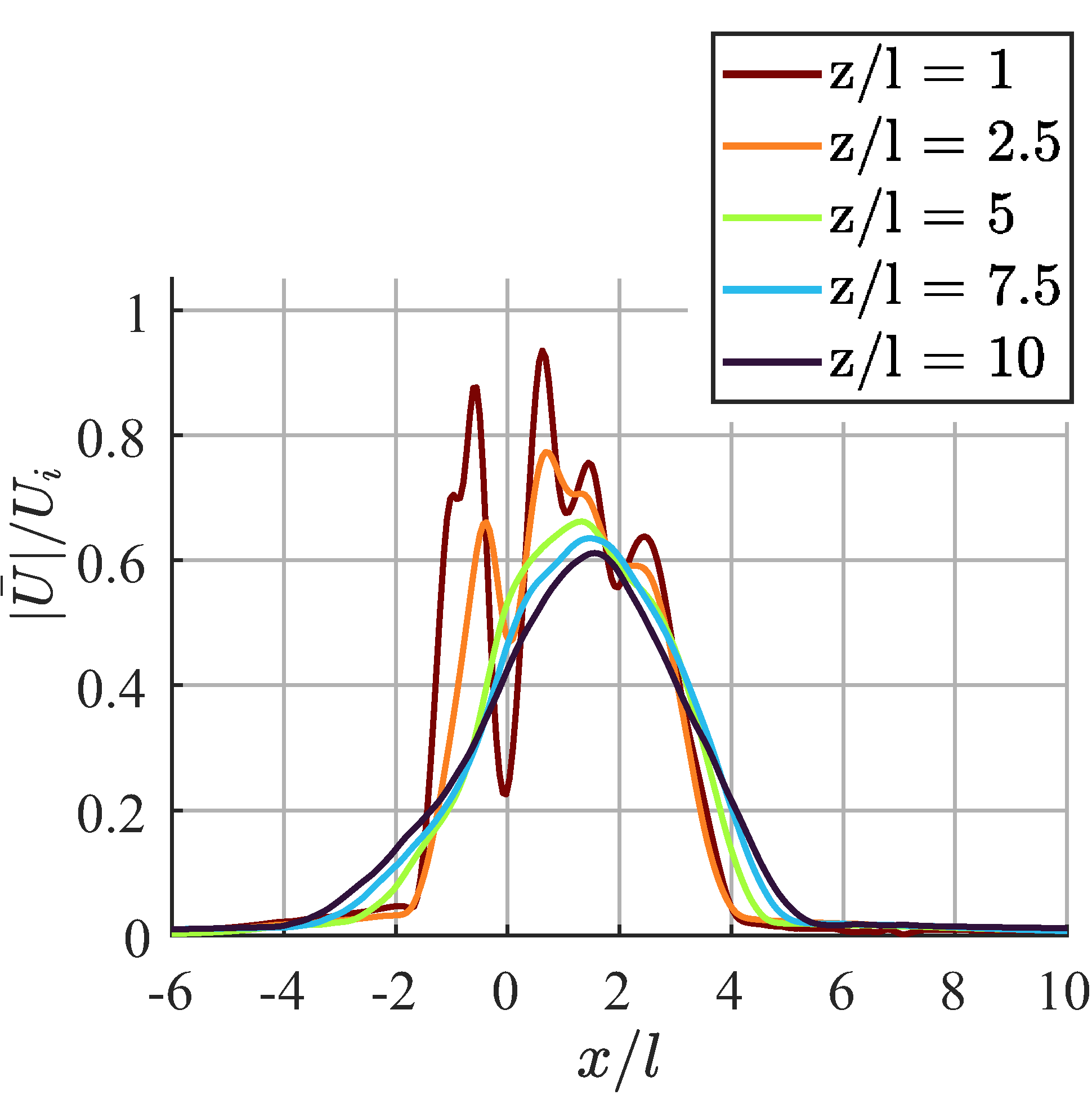}
        \caption{Velocity profile, ${\Delta}z/l = 4$}
        \label{fig:multi_cf_maxM_downwash_speed_D}
    \end{subfigure}
    \caption{Downwash speed ($|\bar{U}|$) of quadrotors in the \emph{offset configuration} (${\Delta}x/l = 2$), along with corresponding velocity profiles below the \emph{lower quadrotor}, normalized by the induced velocity ($U_i$), at various downstream locations ($z/l$), for two vertical separations.}
    \label{fig:multi_cf_maxM}
    \vspace{-0.2in}
\end{figure*}

The vertical separation  
between the downwash-producing upper quadrotor and the lower quadrotor influences the lower neighbor's downwash flow field (cf. Fig.~\ref{fig:multi_cf_maxF_downwash_speed_A} and \ref{fig:multi_cf_maxF_downwash_speed_C}). Larger vertical separation between upper and lower quadrotor causes its wake structure to have more distance to spread out and homogenize, resulting in a wider jet and a smoother inflow by the bottom quadrotor in Fig.~\ref{fig:multi_cf_maxF_downwash_speed_A}. Therefore, an increased vertical separation results in larger magnitudes of velocities below the lower quadrotor as shown in Fig.~\ref{fig:multi_cf_maxF_downwash_speed_B}. Nevertheless, the velocities do not appear similar to the ones below a single quadrotor in Fig.~\ref{fig:single_cf_piv_composite_A} because the lower quadrotor experiences a net downwash flow component due to the presence of the upper quadrotor. 

Beyond the thrust reduction in the lower quadrotor previously illustrated in Fig.~\ref{fig:FM_lower_A}, close vertical separation also results in lower velocity magnitudes (cf. Fig.~\ref{fig:multi_cf_maxF_downwash_speed_B} and~\ref{fig:multi_cf_maxF_downwash_speed_D}). This phenomenon occurs because the lower quadrotor operates within the upper quadrotor's near-field wake---a region characterized by high-velocity, non-uniform flow structures (visible as multiple high velocity lobes across the $x$-axis in Fig.~\ref{fig:multi_cf_maxF_downwash_speed_C}). These complex flow structures directly interfere with the lower quadrotor's ability to generate its own clean downwash, thereby reducing its effective downwash velocity. In contrast, far-field regions ($6.5 < z/l < 10$) resemble flatter, more uniform velocity profiles compared to single quadrotor flow fields, which had unimodal profiles in the far-field (cf. Figs.~\ref{fig:single_cf_piv_composite_B} and~\ref{fig:multi_cf_maxF_downwash_speed_D}). 

Fig.~\ref{fig:multi_cf_maxM} represents case II in Fig.~\ref{fig:case_schematic}, where the upper quadrotor is horizontally offset, resulting in maximum pitch moment on the lower quadrotor. The lower quadrotor no longer receives the strongest part of the upper quadrotor's wake at its center. This change in the inflow condition results in an asymmetry in the lower quadrotor's wake seen in Figs.~\ref{fig:multi_cf_maxM_downwash_speed_A} and~\ref{fig:multi_cf_maxM_downwash_speed_C} at vertical separations of ${\Delta}z/l = 12.5$ and ${\Delta}z/l =  4.0$, respectively. This is evident in the velocity profiles that appear asymmetrical about the axial centerline (cf. Figs.~\ref{fig:multi_cf_maxM_downwash_speed_B} and~\ref{fig:multi_cf_maxM_downwash_speed_D}). The offset configuration results in less direct wake overlap, shifting regions of peak velocity off to one side in Figs.~\ref{fig:multi_cf_maxM_downwash_speed_A} and~\ref{fig:multi_cf_maxM_downwash_speed_C}. This is evident for ${\Delta}z/l \ge 5$, where the peak velocity magnitude is shifted away from the axial centerline (cf. Figs.~\ref{fig:multi_cf_maxM_downwash_speed_B} and \ref{fig:multi_cf_maxM_downwash_speed_D}). 

The lateral offset leads to a left-right imbalance that becomes more pronounced at shorter vertical separations, resulting in an aerodynamic loading on the lower quadrotor causing strong wake interaction and merging below the lower quadrotor as observed in Fig.~\ref{fig:multi_cf_maxM_downwash_speed_C}, whereas for a larger vertical separation in Fig.~\ref{fig:multi_cf_maxM_downwash_speed_A}, the wake below the lower quadrotor blends gradually.

\subsection{Dynamic Quadrotor Interactions}

\begin{figure}[ht!]
    \centering
    \includegraphics[width=0.45\textwidth]{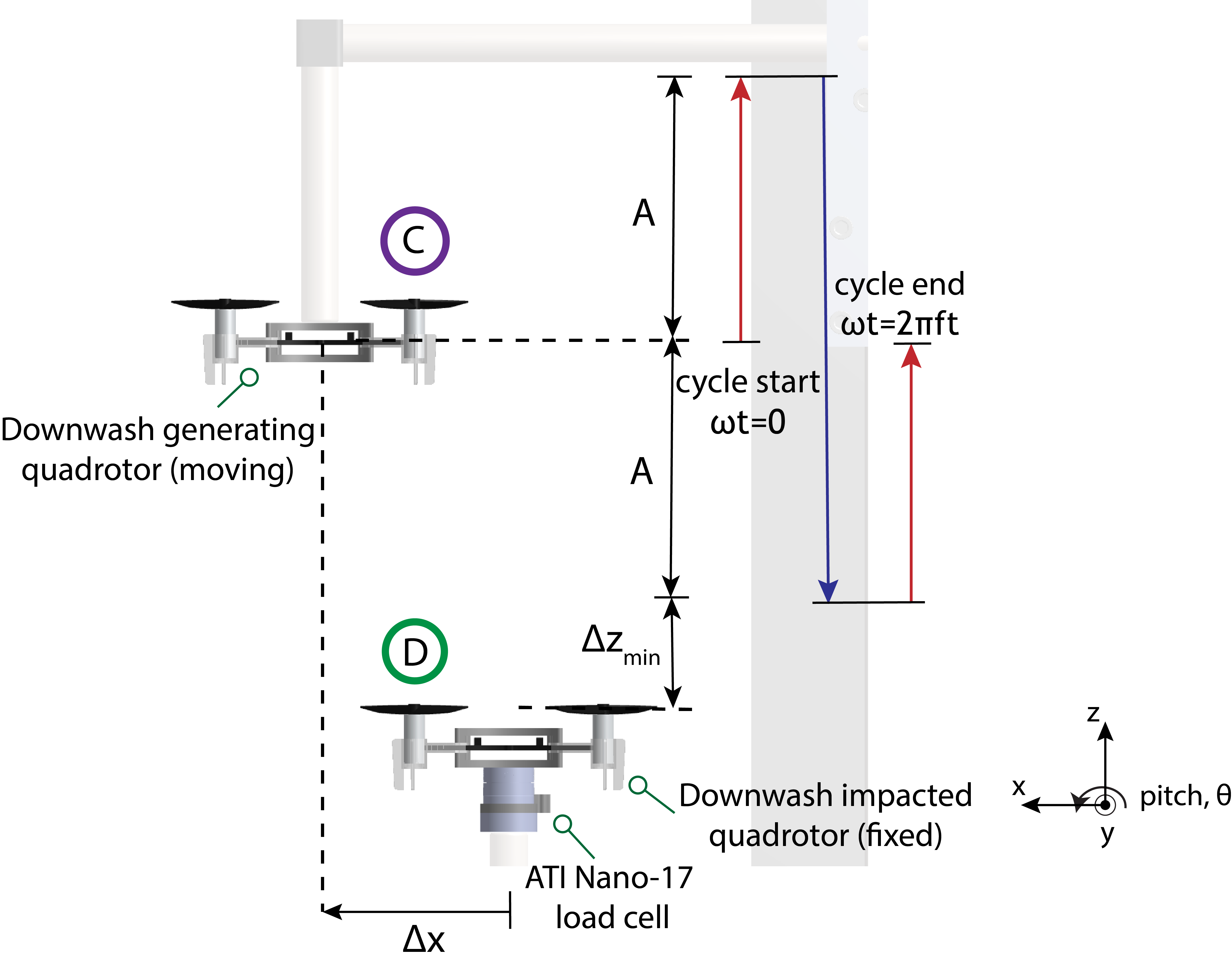}
    \caption{Schematic of dynamic downwash interaction setup with parameters specified in Table~\ref{table:crazyrail_params}. The red arrow indicates the retreating phase, while the blue arrow indicates the approaching phase of the dynamic interaction cycle.
    \label{fig:CrazyRail_schematic}}
\end{figure}

Beyond the static separation effects (${\Delta}x, {\Delta}z$) analyzed in Sections~\ref{subsubsec:F&M_measurements} and~\ref{subsubsec:V_measurements},  
adjacent vehicles experience dynamic interactions that vary with the \emph{rate of change} of separation ($\Delta{\dot{z}}$) as illustrated in Fig.~\ref{fig:cover_image} (right). To characterize these dynamic interactions, we developed a precision linear dynamic acceleration platform, \emph{CrazyRail}, shown in Fig.~\ref{fig:CrazyRail_schematic}, capable of rapidly accelerating a hovering Crazyflie quadrotor (C) towards and away from a stationary hovering Crazyflie quadrotor (D) positioned below. This apparatus enables direct measurement of the interaction effects that emerge during dynamic proximity operations. The stationary lower quadrotor was equipped with an ATI Nano-17 force/torque transducer, enabling measurements in both the stacked configuration and offset configuration ($\Delta x/l = 2$). Data acquisition mirrored those established for the static measurement setup detailed in Section~\ref{subsubsec:F&M_measurements}. 

Quadrotor (C) mounted on the dynamic traverse, follows a prescribed sinusoidal trajectory with three key interaction parameters: oscillation amplitude ($A$), minimum separation distance (${\Delta}z_{min}$) between interaction quadrotors, and motion frequency ($f$), where $f$ relates to the angular frequency ($\omega$) as defined subsequently.
The sinusoidal actuation effectively captures dynamic interaction effects produced by velocity field variations due to changing downwash intensity, while ensuring predictable relative separation between the quadrotors.
\begin{table}[tb]
    \centering
    \begin{tabular}{|l | c | c|}
        \hline
        Parameter & Symbol & Range \\
        \hline\hline
        Minimum $\Delta z$ & $\Delta z_{\min}$ & 0.5\,cm--7\,cm \\
        \hline
        Frequency & $f$ & 0.1\,Hz--1\,Hz \\
        \hline
        Amplitude & $A$ & 2.5\,cm--9\,cm \\
        \hline\hline
        Instantaneous rate of separation & $\Delta{\dot{z}(t)}$ & -0.565\,m/s--0.565\,m/s \\
        \hline
    \end{tabular}
     \caption{\textbf{Parameters for dynamic quadrotor interactions} \label{table:crazyrail_params}}
\end{table}

\begin{figure}[tbh]
    \centering
    \begin{subfigure}[t]{0.235\textwidth}
        \centering
        \includegraphics[width=\linewidth]{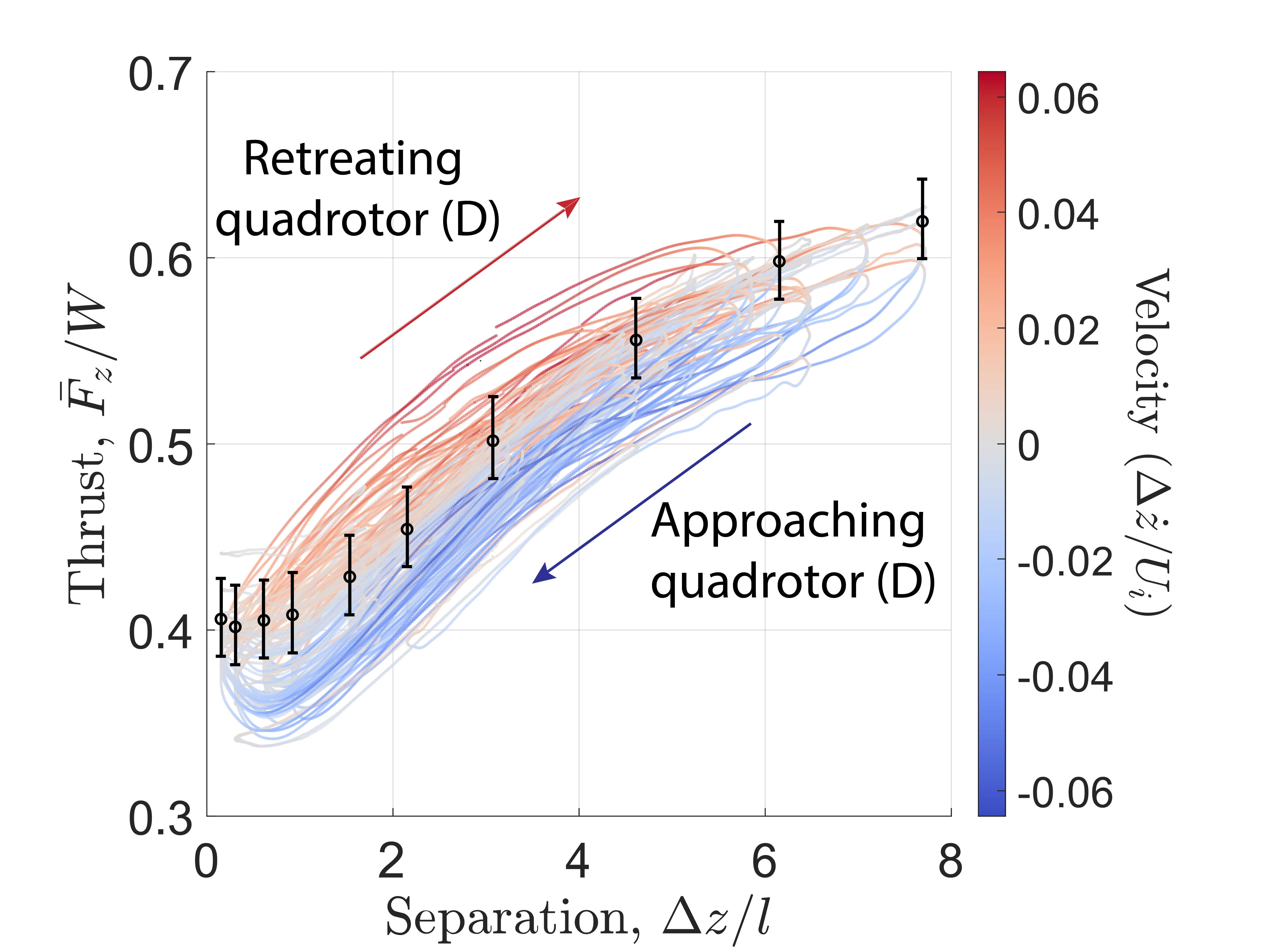}
        \caption{Axial thrust, $\bar{F_z}$}
        \label{fig:cycle_avg_Fz_stacked}
    \end{subfigure}
    \hfill
    \begin{subfigure}[t]{0.235\textwidth}
        \centering
        \includegraphics[width=\linewidth]{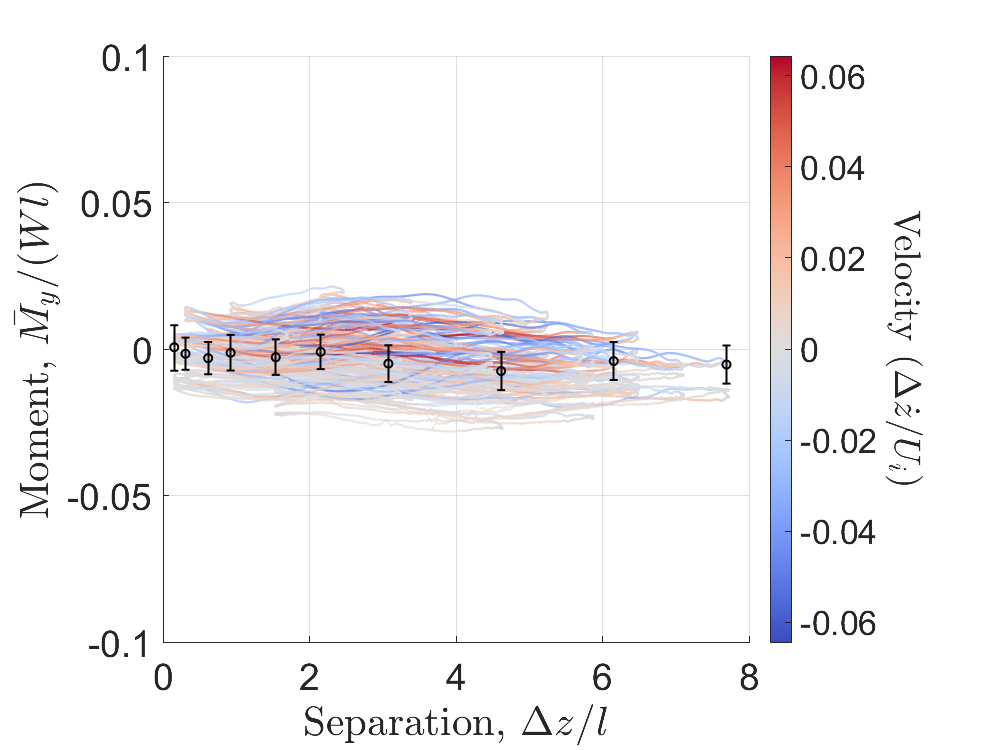}
        \caption{Pitch moment, $\bar{M_y}$}
        \label{fig:cycle_avg_My_stacked}
    \end{subfigure}
    \caption{Cycle-averaged forces and moments of quadrotor (D) in stacked configuration}
    \label{fig:cycle_avg_FM_stacked}
\end{figure}
\begin{figure}[tbh]
    \centering
    \begin{subfigure}[t]{0.235\textwidth}
        \centering
        \includegraphics[width=\linewidth]{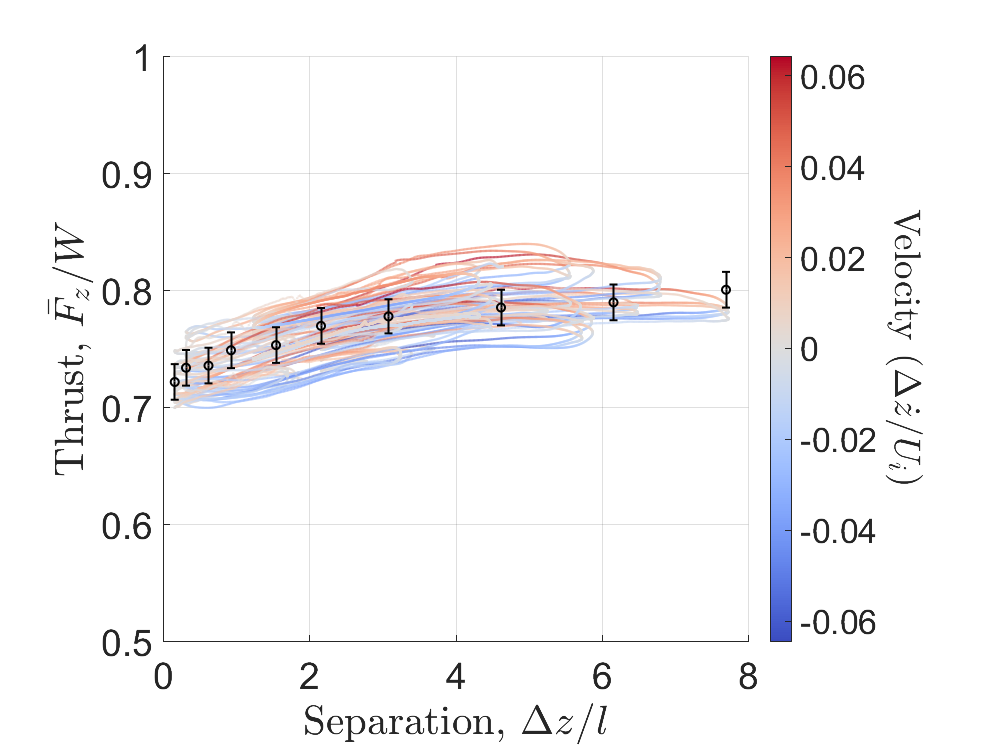}
        \caption{Axial thrust, $\bar{F_z}$}
        \label{fig:cycle_avg_Fz_offset}
    \end{subfigure}
    \hfill
    \begin{subfigure}[t]{0.23\textwidth}
        \centering
        \includegraphics[width=\linewidth]{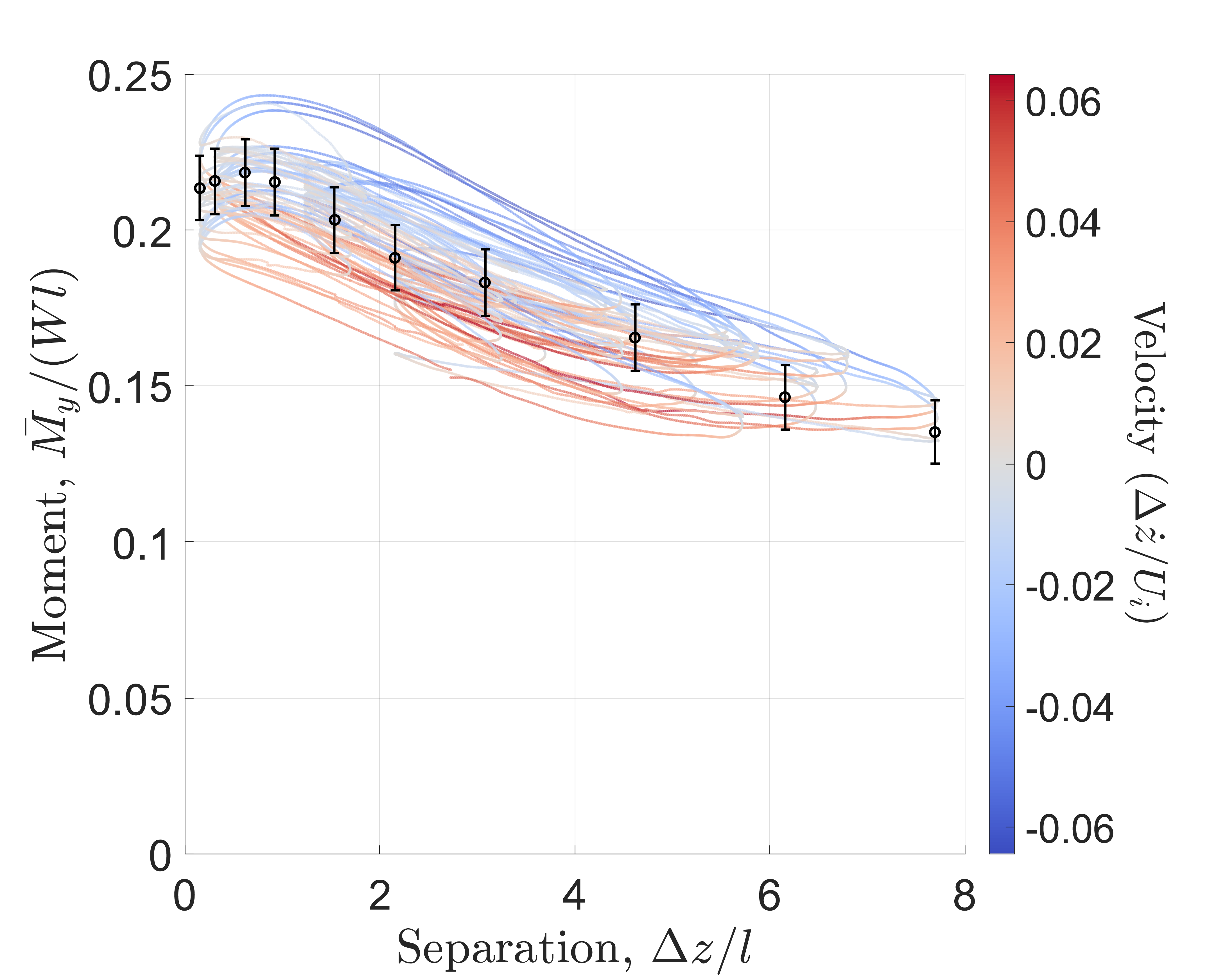}
        \caption{Pitch moment, $\bar{M_y}$}
        \label{fig:cycle_avg_My_offset}
    \end{subfigure}
    \caption{Cycle-averaged forces and moments of quadrotor (D) in offset configuration}
    \label{fig:cycle_avg_FM_offset}
\end{figure}

The separation distance, ${\Delta}z(t)$,  is defined as
\begin{equation}
    \Delta z(t) = \Delta z_{min} + A(1 + \sin(\omega t)),
\label{eqn:unsteady_profile_separation}
\end{equation}
where angular frequency, $\omega=2\pi f$ governs the rate of sinusoidal oscillation.
The rate of change of separation, ${\Delta{\dot{z}(t)}}$, 
is thus $A\omega\cos(\omega t)$.

Each trial captured data through a comprehensive 48-cycle sinusoidal motion sequence. The sequence began with four ramp-up cycles that accelerated the traverse to the target frequency and amplitude. These initial cycles transitioned into forty stable oscillations, forming our primary experimental dataset, capturing the full range of dynamic interactions. This was followed by four ramp-down cycles that provided controlled deceleration, ensuring measurement integrity across the experimental timeline. The transition phases---initial four acceleration cycles and final four deceleration cycles---were removed from the analysis. Force and moments were phase-averaged across these core cycles to generate a time-series data representative of steady-state dynamic interactions, seen as loops in Figs.~\ref{fig:cycle_avg_FM_stacked} and~\ref{fig:cycle_avg_FM_offset} for a given combination of parameters in Table~\ref{table:crazyrail_params}. By overlaying these profiles across the full range of combinations, we can evaluate and compare the influence of separation, ${\Delta}z$, and velocities, ${\Delta{\dot{z}}}$, on the thrust generated by the downwash-impacted quadrotor. 

The blue portion of each loop shows a negative velocity and indicates the approaching phase, when quadrotor (C) reduces its relative separation from quadrotor (D); the red portion indicates a positive velocity and indicates the retreating phase. The loops in Figs.~\ref{fig:cycle_avg_FM_stacked} and~\ref{fig:cycle_avg_FM_offset} appear nearly white at lower frequencies, implying minimal velocity. As the oscillation speed increases, influenced by frequency and separation, the loops broaden, reflecting a greater force and moment variation in quadrotor (D) between the approaching and retreating phases. Simultaneously, the colors transition from white to deeper hues, indicating higher velocities, $\Delta{\dot{z}}$. Black data points and error bars represent mean and standard deviation from static data collected at fixed separations (${\Delta}z$).

Force and moment data were collected for various values of ${\Delta}z_{min}$, ten trials each, forming the standard deviations indicated by the error bars and averaged across those trials at varying separations as indicated by the black circles in Figs.~\ref{fig:cycle_avg_FM_stacked} and~\ref{fig:cycle_avg_FM_offset}. These static points serve as a baseline for force and moment, where there is no relative dynamic motion between quadrotors (C) and (D). The static data presented here aligns with the results in Fig.~\ref{fig:FM_lower_A} and~\ref{fig:FM_lower_C} under Section~\ref{subsubsec:F&M_results}, supporting the observed monotonic decrease in thrust, notably in the stacked configuration (Fig.~\ref{fig:cycle_avg_Fz_stacked}), and the presence of destabilizing moments in the offset configuration (Fig.~\ref{fig:cycle_avg_My_offset}) with decreasing vertical separation.

The effect of dynamic interaction on axial force, $\bar{F}_z$, and pitch moment, $\bar{M}_y$, in the stacked and offset configurations is illustrated in Figs.~\ref{fig:cycle_avg_FM_stacked} and~\ref{fig:cycle_avg_FM_offset}, respectively. The cycle-averaged thrust, $\bar{F}_z$, generated by quadrotor (D) due to dynamic interaction from quadrotor (C) increases with separation (cf Figs.~\ref{fig:cycle_avg_Fz_stacked} and~\ref{fig:cycle_avg_Fz_offset}), where the red arrow indicates the region where quadrotor (C) retreats from quadrotor (D), and the blue arrow indicates the region where quadrotor (C) approaches quadrotor (D). Notably, the approach phase of each interaction cycle has lower thrust in comparison to its retreating counterpart. The stacked configuration demonstrates a wider thrust range across ${\Delta}z$, due to the complete four-motor overlap generating stronger aerodynamic interactions. In contrast, the offset configuration exhibits a narrower (flatter) thrust range with modest changes across ${\Delta}z$, reflecting the partial two-motor overlap that creates distinct approaching and retreating quadrotor behaviors (cf Figs.~\ref{fig:cycle_avg_Fz_stacked} and~\ref{fig:cycle_avg_Fz_offset}). 

The stacked configuration produces negligible pitch moments centered around zero in Fig.~\ref{fig:cycle_avg_My_stacked}, regardless of ${\Delta}z$, indicating balanced aerodynamic interactions between quadrotors. Meanwhile, the offset configuration generated substantial pitch moment (Fig.~\ref{fig:cycle_avg_My_offset}) that decreases in magnitude with increasing separation, reflecting asymmetry across aerodynamic forces requiring control compensation to counter instability. The offset arrangement also demonstrates velocity-dependent behavior, with the approaching phase causing stronger destabilizing moments than the retreating phase. 

\section{Conclusions} 
\label{sec:conclusion}
This paper quantifies the impact of downwash on forces, moments, and velocities by mapping its effects in the near-field and far-field regions beneath quadrotors. Force and moment measurements were captured using six-axis force/torque transducers, and velocity measurements were acquired using PIV. The normalization of these quantities ensures a scalable framework for broader applications.

The force measurements demonstrate that when two quadrotors hover above each other in the stacked configuration at small ranges of vertical separations, the downwash-impacted quadrotor experiences a strong reduction of approximately $65-70\%$ of its hover thrust. In contrast, a horizontal offset of two arm lengths results in instability from a downwash-induced pitch moment, requiring a restoring moment for stability. 

Flow below a single quadrotor in the near field appears as distinct rotor jets with a dead zone between them, and corresponding velocity profiles have multiple peaks. Beyond the merge point ($z/l > 6.5$), these rotor jets have coalesced to form a turbulent jet with self-similar profiles, validating turbulent jet scaling~\cite{PopeTurbulentFlows}, here successfully applied to quadrotors even at this relatively low Reynolds number ($Re \approx 9000$). This is significant because turbulent jet scaling is generally formulated for high Reynolds numbers ($Re \sim 10^4$–$10^5$).

The success of the turbulent jet scaling provides a powerful \textit{reduced-order algebraic model}. The validation of scaling laws for forces, moments (Figs.~\ref{fig:FM_lower} and~\ref{fig:FM_upper}), and velocities (Figs.~\ref{fig:single_cf_PIV}) create a framework defined by four critical parameters: merge point ($z_m/l$), induced velocity ($U_i$), characteristic arm length ($l$), and virtual origin ($z_0$). These scaling relationships apply directly without modification for quadrotors maintaining similar rotor spacing relative to the rotor diameter. Even for vehicles with different rotor spacing relative to rotor diameter and varying propeller counts ($n$), the fundamental self-similarity principles we have validated remain robust, with the four key parameters ($l, z_0, U_i$, and $z_m/l$) determined for the specific vehicle configuration. This consistency aligns with, and expands,  findings from related studies~\cite{BauersfeldRoboticsMeetsFluidDynamics, JainDownwashModel} showing rotor flows merging between 1.5-2.5 motor-to-motor distances across diverse platforms, including custom large quadrotors. By applying these robust merging principles, our analysis extends beyond homogeneous systems to establish a comprehensive framework for heterogeneous multirotor systems.

The flow-field snapshots were extended to characterize and analyze multi-vehicle downwash interactions. For multi-vehicle velocities, the downwash from the upper quadrotor alters the inflow of the lower quadrotor. For a larger vertical separation (${\Delta}z/l = 12.5$) in the stacked configuration (Fig.~\ref{fig:case_schematic}), the downwash jet from the upper quadrotor spreads over a wider space before it reaches the downwash impacted quadrotor, whereas close vertical separation (${\Delta}z/l = 4$) in such configuration results in a concentrated wake that interacts strongly with the lower quadrotor, reducing velocity magnitudes, as well as thrust generated by the downwash-impacted, lower quadrotor. Meanwhile, for the offset configuration in Fig.~\ref{fig:case_schematic} (right), the presence of a horizontal offset (${\Delta}x/l = 2$) resulted in a shift in the strongest velocity regions away from the lower quadrotor's axial centerline, altering the symmetry of merged wake below the downwash affected quadrotor.

Pronounced aerodynamic interference is measured when quadrotors interact dynamically, with peak disturbance magnitudes documented during stacked vertical approaches between quadrotors. In contrast, the retreat phases result in weaker interaction magnitudes, permitting the impacted quadrotor to attain stable airflow and recover its thrust. Valuable insights into the separation rate (${\Delta{\dot{z}}}$) and its corresponding influence on downwash intensity are revealed through these dynamic interactions. A comprehensive aerodynamic dataset is established through these measurements, enabling the development of flow-aware controllers capable of anticipating and counteracting dynamic load variations encountered during multi-vehicle interaction. In the stacked configuration, dynamic approaches and retreats generate symmetrical force patterns that balance out moments. In contrast, the offset configuration induces destabilizing pitch moments on the downwash-impacted quadrotor, with higher magnitudes during the approach phase than the retreat dynamics.

\section{Limitations and Future Work}
\label{sec:Limitations}
The experiments and analyses outlined in this paper were conducted using Crazyflie quadrotors, focusing on homogeneous configurations, i.e., interactions between identical small-small quadrotors with the same rotor configuration, size, and mass. While this study examines uniform setups, these findings can be extended to more diverse multi-rotor systems. Notably, near-field wake interactions and merging behavior can differ in asymmetric configurations, such as heterogeneous multi-rotor systems with differing numbers of rotors and layouts. However, in the far field, once the rotor wakes have merged, the flow transitions to a turbulent jet, where the same fundamental scaling laws validated in this paper apply regardless of the rotor shape or configuration. Future work will explore these near-field variations and their implications for aerodynamic interaction in heterogeneous multi-rotor systems. 

Our experiments on dynamic quadrotor interactions were constrained by the length of traverse (Fig.~\ref{fig:cover_image} right), restricting it to a range of separations ($0.15 \le {\Delta}z/l \le 7.5$). Future work will consider the exploration of larger separations and varying configurations~\cite{KiranConfigurations} for dynamic interactions between quadrotors for a full range of thrust recovery, similar to a larger range of separations as outlined in the static force and moment experiments in Section~\ref{subsubsec:F&M_results}.

\section{Acknowledgments}

This work was supported by a Brown University Seed Award from the Office of the Vice President for Research.  AK is supported by an NSF Graduate Research Fellowship (Award 2040433). The authors gratefully acknowledge the assistance of Alexander Wang for his help in the setup and data collection. 

\bibliographystyle{plainnat}
\bibliography{references}

\end{document}